\documentclass[12pt]{article}
\usepackage[utf8]{inputenc}

\usepackage[T1]{fontenc}
\usepackage[english]{babel}
\usepackage{lmodern, csquotes, eurosym}
\usepackage[style=authoryear,texencoding=utf8,backend=biber,mincitenames=1,maxcitenames=3,maxbibnames=99]{biblatex}
\usepackage[pdfusetitle, colorlinks=true]{hyperref}
\usepackage{caption}
\usepackage{comment}
\usepackage{placeins}
\usepackage{subcaption}
\usepackage{mathtools}
\usepackage{multirow}
\usepackage{dsfont}
\usepackage{amsmath,bm}
\usepackage{booktabs}
\usepackage{array}
\usepackage{multirow}
\usepackage{colortbl}

\usepackage{amssymb}
\usepackage{amsfonts}
\usepackage{geometry}
\usepackage{graphicx}
\usepackage[table]{xcolor}
\usepackage{placeins}

\renewcommand{\baselinestretch}{1.15}
\renewbibmacro{in:}{%
  \ifentrytype{article}
    {}
    {\bibstring{in}%
     \printunit{\intitlepunct}}}

\title{Human Responses to Algorithmic Correction:\\ Evidence from Centre Court
\thanks{We thank Lucas Lippman, Anastasiia Morozova, and Leshan Xu for their excellent research assistance. We are very grateful to the executives at Hawk-Eye Innovations Ltd and a leading tennis official at the time of Hawk-Eye's introduction for providing valuable institutional information. We also thank David Argente, Meghan Busse, Colin Camerer, Andrew Caplin, Ben Enke, George Georgiadis, Mitchell Hoffman, Alex Imas, John List, Christopher Mills, Devin Pope, Michael Powell, Ashesh Rambachan, Antonio Rangel, Paul Raschky, Raphaël Raux, Luis Rayo, Yuval Salant, Jörg Spenkuch, Andrew Wooders, and audiences at EC 2024, BDRM, AI and the Future of Work at Wharton, ESIF Economics and AI+ML Meeting, Machine Learning in Economics Summer Conference, Zurich Workshop in AI+Economics, WEBEAS, MIT Sloan, Monash, UT Dallas, UH Manoa, Kellogg Strategy Brown Bag Seminar, and Booth Behavioral Lab for many helpful comments. We thank the Sloan Foundation for supporting this research under the ``Cognitive Economics at Work'' grant.}
}
\author{David Almog\thanks{Harvard Business School,  \href{mailto:dalmog@hbs.edu}{dalmog@hbs.edu}.}\:, Romain Gauriot\thanks{Deakin University,  \href{mailto:romain.gauriot@deakin.edu.au}{romain.gauriot@deakin.edu.au}.}\:,
Lionel Page\thanks{University of Queensland, \href{mailto:lionel.page@uq.edu.au}{lionel.page@uq.edu.au}.}\:, and Daniel Martin\thanks{University of California, Santa Barbara, \href{mailto:danielmartin@ucsb.edu}{danielmartin@ucsb.edu}.}}
\hypersetup{pdfauthor={David Almog, Romain Gauriot, Lionel Page, and Daniel Martin}}
\date{November 4, 2026}

\begin{document}
\begingroup
\renewcommand{\baselinestretch}{1}\selectfont
\setlength{\parskip}{0pt}
\maketitle

\begin{comment}
\begin{abstract}
We study how algorithmic review changes the initial human decisions it is designed to correct. Using Hawk-Eye ball-location records from professional tennis, we compare officials’ initial calls with and without player-initiated review against a common accuracy benchmark. Our central finding is that officials were more likely to call marginal balls in under review. Among balls landing within 20 mm outside the line, the estimated rate of mistaken “in” calls was 8.6 percentage points higher than without review. Among balls within 20 mm inside the line, the rate of mistaken “out” calls was 4.1 percentage points lower. This shift is consistent with avoiding mistaken “out” calls, whose reversal can require an additional, potentially contentious judgment. Within matches, subsequent close calls shift toward avoiding the type of mistake just overturned, and such adjustment is larger after a mistaken “out” call. For calls within 100 mm of the line, we find no evidence that overall initial accuracy deteriorated under review, despite concerns that workers may rely on correction as a safety net. We derive conditions under which an organization should invest in review technology, incorporating human responses alongside the benefits and costs of subsequent correction. Holding human decisions fixed can therefore misstate the returns to adoption.
\end{abstract}
\end{comment}

\begin{abstract}
We study how algorithmic review changes error composition using Hawk-Eye records from professional tennis. Within 20 mm of the line, mistaken “in” call rates increase by 8.6 percentage points, while mistaken “out” call rates decrease by 4.1 points. Within matches, calls shift toward avoiding previously overturned errors. We find no evidence that initial accuracy deteriorates within 100 mm of the line, despite concerns that workers may rely on correction as a safety net. We derive an investment framework incorporating human responses alongside correction benefits and remedy costs, showing that holding human decisions fixed can misstate the organizational returns to review.
\end{abstract}

\endgroup
%%Long Abstract
%%Organizations increasingly deploy algorithms not only to advise workers but also to review and correct their decisions. Because workers anticipate review, system performance depends on how their decisions adjust before the algorithm intervenes. We study this response using Hawk-Eye in professional tennis, where umpires make repeated binary decisions and the system provides a highly accurate benchmark for correctness. After review was introduced, umpires became more likely to call marginal balls in. Initial-call errors were 8.6 percentage points higher for balls just outside the line and lower for comparably close balls inside. Despite concerns that workers may rely on review as a safety net, we find no evidence that average initial-call accuracy deteriorated. Within matches, reversals were followed by calls in the direction of the correction, with larger adjustments than after challenges that upheld the initial call, consistent with adaptation to being overruled. The results show that evaluating algorithmic review requires accounting for workers’ endogenous responses, especially when corrections are visible and different errors carry different expected consequences under review.

% \begin{quote}
% JEL Codes: D81, D91, O33
% Ashesh: C10, C55, D81, D84

% Keywords: Artificial intelligence, machine learning, behavioral economics, oversight
% \end{quote}

\newpage 
\section{Introduction}
Many organizations now use algorithmic systems to advise workers.\footnote{For example, see \textcite{Hoffman2018,Kawaguchi2020,Agarwal2023,Mills2022}.} Another use is algorithmic review, in which a system checks human decisions and can block or reverse likely mistakes. Algorithmic screening of human pricing interventions provides one example (\cite{Huelden2024}). Related work in law and medicine studies broader uses of algorithms to evaluate human decisions and allocate cases between humans and machines (\cite{Kleinberg2018,Rambachan2022,Raghu2019}). The prospect of review may itself change how workers decide. Correction may reduce the perceived need for care, while public scrutiny may increase it. Review may also change the relative costs of different errors to workers. If review makes certain mistakes more likely to be exposed or more costly to reverse, workers may favor decisions that avoid those mistakes, even at the risk of making others. The value of review therefore depends on both the mistakes it corrects and the changes it induces in initial human decisions.

We study the introduction of Hawk-Eye review in professional tennis, beginning in 2006. Officials make initial calls about whether a ball landed in or out without seeing the system’s assessment. Players can challenge a call and obtain algorithmic review. Hawk-Eye uses camera data to reconstruct where the ball landed, and its assessment is displayed in the stadium. If it contradicts the initial call, that call is publicly overturned. Because players initiate review, some mistaken calls remain unchallenged and uncorrected. Officials nevertheless make their initial decisions knowing that a challenge may lead to a public reversal. The protocol also creates an important asymmetry: the remedy after a successful challenge depends on the initial call. During a rally, if officials initially call a ball “in” but Hawk-Eye shows that it landed out, the reversal directly establishes who won the point. If officials initially call a ball “out” but Hawk-Eye shows that it landed in, the initial call has stopped play. The chair umpire must then decide whether to award the point or replay it, judging whether the player facing the shot could have returned it had play continued. That additional discretionary ruling can itself be disputed, potentially making reversals of initial “out” calls more contentious for officials.

Hawk-Eye’s introduction provides an unusually clean setting for measuring human decisions with and without algorithmic review. Pretrial release illustrates several challenges that also arise when evaluating human decisions in settings such as medical screening and lending. Classifying a release decision as a mistake requires specifying how the risks of release are weighed against the costs of detention (\cite{Kleinberg2018}). These decisions also generate selective labels: researchers cannot observe whether detained defendants would have returned to court had they been released (\cite{Lakkaraju2017,Rambachan2022}). Judges may also possess information unavailable to the algorithm, so disagreement need not indicate a human error. Tennis provides a clearly defined accuracy criterion—where the ball landed—with little scope for private information to justify a different classification. Hawk-Eye records ball locations regardless of officials’ calls and players’ challenges, providing a highly accurate benchmark for both challenged and unchallenged decisions. Because it recorded locations at some tournaments before players could challenge calls, we can apply the same benchmark under both regimes.\footnote{Before stadium review, Hawk-Eye replays could already inform television commentary and expose incorrect calls to public criticism. The review regime introduced player challenges and immediate, public reversal of official decisions during the match.} A further difficulty in studying review is that researchers may observe only decisions after correction. Combining Hawk-Eye location records, challenge records, and video audits allows us to reconstruct officials’ initial calls. These features allow us to measure both the overall rate and composition of initial errors, separately from subsequent correction.

Our central finding is that officials were more likely to call marginal balls in under review, with more mistaken “in” calls and fewer mistaken “out” calls near the boundary. We compare initial calls under the two regimes, accounting for ball location, speed, score, and match characteristics. Among balls landing within 20 mm outside the line, the estimated rate of mistaken “in” calls was 8.6 percentage points higher than without review. Among balls within 20 mm inside the line, the rate of mistaken “out” calls was 4.1 percentage points lower. This pattern is consistent with officials placing greater weight on avoiding the error whose reversal can require a contentious ruling about whether to award the point or replay it.

Several pieces of evidence support interpreting this shift as a response to review. Ball-location and speed distributions are similar across regimes, and the central pattern persists with richer controls for shot and match characteristics. Because adoption was not randomized, we also restrict the comparison to named events observed under both regimes and include named-event fixed effects. The estimated reallocation of errors remains economically sizable, although less precise. Institutional information also indicates no substantial changes to umpire training and performance assessment, the pool of umpires, or the positioning and instructions given to line judges.\footnote{These institutional details were provided by a leading tennis official at the time of Hawk-Eye’s introduction.}

Comparisons within matches provide additional evidence that officials adjust their calls following public correction, without relying on differences across review regimes. After a reversal, officials’ subsequent close calls shift in the direction that would avoid repeating the same type of mistake. This adjustment is stronger after a mistaken “out” call, consistent with officials being especially reluctant to repeat a mistake that could provoke further dispute.

A separate concern about adopting review is that workers may take less care if they expect the system to catch their mistakes. We find no evidence of a deterioration in overall initial accuracy. For balls landing within 100 mm of the line, the estimated initial error rate was 1.1 percentage points lower with review than without it (SE = 0.8 percentage points), although this difference was not statistically significant. For non-serves, the estimated reduction was 2.3 percentage points (SE = 1.2 percentage points), significant at the 10\% level, while the estimate for serves was close to zero. One possible explanation is that the prospect of public reversal encourages greater care, with slower non-serves leaving more scope for attention to improve accuracy.

We then examine how these differences in human decisions affect the organizational returns to algorithmic review. A calculation that applies the technology to a fixed set of human decisions captures the benefits of correction but misses changes in the mistakes workers make. We compare this benchmark with a calculation incorporating the initial error rates observed under review, holding correction fractions and unit costs fixed. Both calculations account for losses from mistakes that remain uncorrected and the costs of remedying corrected mistakes. This distinction matters because restoring the correct classification need not undo the consequences of the original decision, such as interrupting a rally that should have continued.

Calibrating these calculations to the observed error and correction rates shows that accounting for human responses can either strengthen or weaken the case for adoption. When remaining errors carry equal losses and remedies are costless, incorporating the observed initial decisions supports approximately 32\% more expenditure on review than the fixed-decision benchmark. The ordering reverses when the loss from a remaining mistaken “in” call exceeds roughly four times the loss from a remaining mistaken “out” call. Under those same assumptions about remedies, the increased weight on mistaken “in” calls outweighs the benefit of fewer mistaken “out” calls. Thus, even when fewer errors remain, their composition can make review less valuable than a calculation holding human decisions fixed would suggest.

\begin{comment}
Our findings speak most directly to selective review systems in which workers make initial decisions, corrections are visible, and reversing different errors carries different consequences.\footnote{Appendix \ref{sec:ev} assesses the external validity of our findings using the framework of \textcite{List2020}.} Similar asymmetries can arise elsewhere: lifting an erroneous block on a payment and recovering a payment that should have been blocked involve different remedies. The visibility and personal consequences of correction may differ across applications. Extending the evidence therefore requires examining who initiates review, how corrections are communicated, and which consequences decision makers bear.
\end{comment}

The next subsection relates our contribution to the literature. Section \ref{sec:settingdata} describes the setting and data. Section \ref{sec:results} presents the evidence on error reallocation, discusses identification, and examines responses to prior reversals and overall initial accuracy. Section \ref{sec:hetero} explores heterogeneity across tournament characteristics and between serves and non-serves. Section \ref{sec:review_returns} develops the investment framework and uses the observed error and correction rates to illustrate when accounting for human responses changes the case for adoption. Section \ref{sec:conclusion} concludes.

\subsection{Related Literature}\label{sec:lit}

Our evaluation framework is closely related to work on the design of human--AI collaboration. \textcite{AgarwalMoehringWolitzky2025} characterize collaboration policies using human accuracy responses under assumptions that allow these responses to predict performance across policies. \textcite{Agarwal2023} evaluate collaboration between radiologists and AI using diagnostic accuracy and the cost of human time. Both papers relate human behavior to the objective of a designer choosing how to use AI. We study review after the human decision, where correcting a classification need not eliminate the costs of the initial mistake. Our framework relates the observed human response to the quantities needed for evaluation.

This paper contributes to the literature on human--algorithm interaction by studying ex post algorithmic review. Most empirical work examines systems that provide recommendations before a human decision. This literature shows that system performance depends on how decision makers use algorithmic advice. Discretion may reflect biases rather than private information \parencite{Hoffman2018}. Adherence varies across workers and decision contexts and increases when workers' forecasts are incorporated into the algorithm \parencite{Kawaguchi2020}. Risk scores may also reallocate attention rather than induce mechanical compliance \parencite{Mills2022}. Belief updating from qualitative AI recommendations depends systematically on prior beliefs and on whether those recommendations confirm or contradict them \parencite{Kovach2026}. Miscalibrated beliefs can produce systematic overreliance or underreliance on algorithmic advice \parencite{Caplin2025}, while decision makers may adjust their advice weights too little as the value of their private information changes \parencite{Balakrishnan2026}. Accordingly, providing AI predictions need not improve average performance \parencite{Agarwal2023}, while appropriately coarsening the information conveyed by AI risk scores can improve decisions \parencite{HoongDreyfuss2026}. These studies examine how the information supplied by an algorithm shapes assisted decisions. Our setting instead removes the contemporaneous recommendation. Line judges make their initial calls without observing the system's assessment. The relevant behavioral margin is therefore how the prospect of subsequent review changes their decision rule before the algorithm intervenes.

Algorithmic review lies between recommendations and full automation. Work on human--algorithm decision rights considers how authority should be allocated \parencite{Athey2020,Ide2025} and how predictions can support automation, triage, or oversight \parencite{Kleinberg2018,Raghu2019,Rambachan2022,Huelden2024}. Under review, the human acts first but an algorithm can subsequently alter the outcome. The review protocol thus affects both which mistakes are corrected and the consequences attached to different initial decisions. Algorithmic systems can therefore shape choices not only by conveying information but also by changing action costs. \textcite{McLaughlin2024} model recommendations as defaults that make deviations costly, while \textcite{Albright2024} finds that lenient bail recommendations affect judges' decisions even when the underlying risk information is unchanged, consistent with reputational cover. Our findings are consistent with analogous incentive effects under ex post review, where correction changes the consequences attached to different errors. Accurate review can consequently shift the initial decision threshold and reallocate errors across types, so system performance cannot be inferred from a fixed distribution of human decisions. On a different dynamic margin, \textcite{Cho} examines skill depreciation following automated assistance among baseball umpires. More recent work on MLB's 2026 Automated Ball-Strike challenge system shows that asymmetric challenge pressure can shift umpires' decision thresholds within games \parencite{Michopoulos2026} and documents brief adjustments near the decision boundary following overturns and players' decisions to invoke review \parencite{Lee2026}. These findings provide evidence from a subsequent setting consistent with our central result that selective algorithmic review is associated with differences in initial human decisions.

Related evidence shows that people respond differently to human and algorithmic assessment. \textcite{Avery2023} show that anticipated assessment by AI rather than humans changes job seekers' application behavior, while \textcite{AlmogLippmanMartin2026} find that workers alter the quantity and quality of their output when evaluated by AI. Our setting does not allow us to distinguish a response specific to algorithmic review from a response to the introduction of review more generally. At the same time, it illustrates why algorithms can expand the scope of review. Hawk-Eye makes it possible to assess and correct decisions in real time with a combination of speed and accuracy that human review could not feasibly match.

Anticipated review also connects our paper to the literature on monitoring and accountability. Monitoring can change opportunistic behavior, compliance, and worker productivity \parencite{Nagin2002,Olken2007,Gosnell2020,Zou2021}. A distinctive feature of algorithmic review is that monitoring can both expose an error and remedy it in real time. These functions create potentially offsetting incentives. The availability of correction may provide a safety net, while disruptive or consequential reversals may induce workers to avoid the decisions most likely to be overturned. Moreover, when the likelihood or consequences of correction differ across errors, workers may shift their decision thresholds rather than simply exert more or less effort. Our evidence documents differences in error composition under review that aggregate accuracy alone does not capture.

Finally, the visibility of reversals connects to work on social image and observability \parencite{Bursztyn2017}. Experimental evidence shows that image concerns can lead decision makers to disregard AI recommendations even when others’ perceptions carry no monetary consequences \parencite{AlmogNIIC}, and that making workers’ reliance on AI visible to an evaluator reduces adoption and performance \parencite{AlmogJMP}. Our setting combines public visibility with reversal and remediation, whereas these experiments isolate responses to observability.

\section{Setting and Data}\label{sec:settingdata}

Hawk-Eye review was first used at an ATP Tour event in March 2006 at the Nasdaq-100 Open in Key Biscayne (currently known as the Miami Open).\footnote{The ATP Tour is the top tennis tour organized by the Association of Tennis Professionals.} Later that year, the US Open and many other tournaments on surfaces other than clay adopted this new technology by allowing players to challenge calls.\footnote{Players have a fixed number of challenges available during each set, but successful challenges do not count against this limit. Players were not allowed to challenge chair umpire decisions on non-clay courts before this system was put in place.} Hawk-Eye uses six to ten computer-linked television cameras positioned around the court to collectively create a three-dimensional representation of the ball's trajectory. Hawk-Eye reports having an average error of just 3.6 mm, so just like the ATP Tour we will treat Hawk-Eye readings as if they are the ground truth.\footnote{For perspective, the standard diameter of a tennis ball is 67 mm.}
% Add back in revisions:
% \footnote{See \textcite{Palacios2023} for a comprehensive review of other sports settings that capitalized on favorable empirical situations to address economic questions.}

\subsection{Professional Tennis, Umpires, and Hawk-Eye}

Professional tennis is a racket sport that is played either one-on-one (singles) or two-on-two (doubles). In singles tennis, two players compete against each other on opposite sides of the tennis court. The objective is to score points by hitting the ball ``in'' (within the bounds of the opponent's side of the court). We will concentrate solely on men's singles matches (singles matches where both players are men) because the analysis of other formats is underpowered in our data.\footnote{If more data were available on the other formats, it might be of interest to examine whether responses to algorithmic review vary by gender or team composition.} The scoring system for tennis is based on a series of points, games, and sets. Players accumulate points to win a game and accumulate games to win a set. Matches are typically played as the best of three or five sets, with each set requiring a player to win at least six games.

A crew of up to ten umpires is involved in officiating a match. The roles typically include one chair umpire who oversees the entire match, making decisions on points, penalties, and overall match control. Additionally, there are usually nine line umpires positioned around the court, each responsible for specific lines, with the sole duty to determine whether the ball bounced in or out when it is relatively close to their respective line.\footnote{Figure \ref{fig:Umpires} provides a visual representation of the positioning of each umpire.} Because chair umpires can overrule line judges and the data do not identify individual officials, we evaluate crew-level decisions rather than individual performance or within-official adjustment.

International chair umpires are certified with Gold, Silver, or Bronze badges, while line umpires are graded according to national or other systems. Certification and grading methods remained unchanged during the study period, and Hawk-Eye metrics were not incorporated into these evaluations for the first couple of years. A leading tennis official who was involved in testing and introducing Hawk-Eye provided valuable institutional insights that will be used throughout the paper.

The Hawk-Eye review protocol works by endowing players with 2-3 challenges per set.\footnote{In practice, players very rarely exhaust their challenges. This is advantageous for our research question, as umpires are almost always under the threat of being challenged.} If a challenge is successful, players do not lose a challenge opportunity. When a player challenges a call, a computerized path and the final landing location of the ball are displayed on a large screen in the stadium for the umpire, players, and the crowd to observe the outcome of the challenge.

The remedies differ according to the initial call. Reversing an incorrect in call establishes that the ball was out and ordinarily awards the point to the challenger. A first serve is an exception: a fault ordinarily gives the server a second serve, rather than ending the point. When an incorrect out call is reversed, the chair umpire must judge whether the call hindered play before awarding or replaying the point.\footnote{See the \href{https://www.itftennis.com/media/7221/2026-rules-of-tennis-english.pdf}{ITF Rules of Tennis}, Rules 20 and 24 and Appendix VII, Case 7 and Electronic Review Procedures.} This is discretion exercised under the rules. The reversal corrects the line classification, but an interrupted rally cannot resume from the position it would have reached without the call. A replay therefore need not restore the outcome of uninterrupted play.

\subsection{Data}

We combine Hawk-Eye records of ball locations from over 100 tournaments, challenge records, and video audits to reconstruct initial umpire calls.\footnote{We excluded from the analysis the 15 clay court tournaments. We elaborate on this decision in Appendix \ref{AppendixA1}.} The location records provide the accuracy benchmark, challenge records identify reversals, and video audits guide and assess the rules for matching records and classifying calls. We were able to merge 2,038 out of the 2,108 challenges registered for the 28 tournaments common to the two data sets (a 97\% merge rate).

The consolidated data set produced by merging the Hawk-Eye Base data set and the Challenge data set comprises a total of 698 matches across 35 tournament editions, and summary statistics for this data set can be found in Table \ref{tab:SumStats}.\footnote{Table \ref{tab:Tournaments} provides information on the month, category, court type, and the number of matches for each tournament.} This data set includes 109 matches from seven tournament editions that were played before Hawk-Eye review and 589 matches from 28 tournament editions after Hawk-Eye review was active. The samples before and after Hawk-Eye review are very similar in the characteristics of points most relevant for identification, especially the shares of bounces within 20 mm and 100 mm of the line.  The differences in tournament and round composition motivate the inclusion of match-level controls in the analyses below. Figure \ref{fig:PlayerStrokes} shows that the distribution of players’ strokes was highly similar before and after the introduction of Hawk-Eye review, both in proximity to the line and in speed.

\begin{table}[htbp]
\centering
\scalebox{0.79}{
\begin{tabular}{l c c c c}
\hline
\hline
& \multicolumn{1}{c}{Before Hawk-Eye review} & \multicolumn{1}{c}{After Hawk-Eye review} & \multicolumn{1}{c}{Total} & \multicolumn{1}{c}{Post - Pre} \\
& \multicolumn{1}{c}{(PostHK=0)} & \multicolumn{1}{c}{(PostHK=1)} &  & \multicolumn{1}{c}{[Std. diff.]} \\
\hline

\multicolumn{5}{l}{\textit{A. Sample counts}} \\
\hspace{4mm}Players       & 71      & 161     & 174     &  \\
\hspace{4mm}Tournament editions   & 7       & 28      & 35      &  \\
\hspace{4mm}Matches       & 109     & 589     & 698     &  \\
\hspace{4mm}Points        & 15,435  & 83,902  & 99,337  &  \\
\hspace{4mm}Bounces       & 71,630  & 402,467  & 474,097  &  \\
[0.6em]

\multicolumn{5}{l}{\textit{B. Tournament tier (share of tournaments)}} \\
\hspace{4mm}ATP 250       & 28.6\%  & 42.9\%  & 40.0\%  & +14.3 p.p. [0.30] \\
\hspace{4mm}ATP 500       & 0.0\%   & 10.7\%  & 8.6\%   & +10.7 p.p. [0.49] \\
\hspace{4mm}ATP 1000      & 71.4\%  & 46.4\%  & 51.4\%  & -25.0 p.p. [0.53] \\
[0.6em]

\multicolumn{5}{l}{\textit{C. Match round (share of matches)}} \\
\hspace{4mm}Final         & 6.4\%   & 4.6\%   & 4.9\%   & -1.8 p.p. [0.08] \\
\hspace{4mm}Semifinal     & 12.8\%  & 9.0\%   & 9.6\%   & -3.8 p.p. [0.12] \\
\hspace{4mm}Quarterfinal  & 17.4\%  & 14.9\%  & 15.3\%  & -2.5 p.p. [0.07] \\
\hspace{4mm}Round of 16   & 12.8\%  & 19.2\%  & 18.2\%  & +6.3 p.p. [0.17] \\
\hspace{4mm}Other         & 50.5\%  & 52.3\%  & 52.0\%  & +1.8 p.p. [0.04] \\
[0.6em]

\multicolumn{5}{l}{\textit{D. Bounce composition (share of all bounces)}} \\
\hspace{4mm}$<20$ mm from the line   & 0.78\% & 0.69\% & 0.70\% & -0.09 p.p. [0.01] \\
\hspace{4mm}$<100$ mm from the line  & 3.66\% & 3.53\% & 3.55\% & -0.13 p.p.   [0.01] \\
\hspace{4mm}Serves                   & 29.2\% & 27.6\% & 27.8\% & -1.64 p.p. [0.04] \\
\hspace{4mm}Non-serves               & 70.8\% & 72.4\% & 72.2\% & +1.64 p.p. [0.04] \\
[0.6em]

\multicolumn{5}{l}{\textit{E. Average speed in km/h (s.d.)}} \\
\hspace{4mm}Serves      & 147.7 (24.2) & 150.0 (23.1) & 149.6 (23.3) & +2.3 [0.10] \\
\hspace{4mm}Non-serves  & 81.7 (21.4)  & 83.3 (20.4)  & 83.1 (20.5)  & +1.6 [0.08] \\
\hspace{4mm}All         & 101.0 (37.4) & 101.7 (36.5) & 101.6 (36.6) & +0.7 [0.02] \\
\hline
\hline
\end{tabular}
}
\caption{Summary statistics for the consolidated data set. The last column reports the difference in the sample characteristics before and after Hawk-Eye review. Bracketed values are absolute standardized differences.}
\label{tab:SumStats}
\end{table}

\subsubsection{Hawk-Eye Base Data Set}

The main source of data for this paper is the Hawk-Eye Base data set, which consists of the official Hawk-Eye data for matches played at the international professional level between March 2005 and March 2009. This data set provides a number of pieces of information for each point, including the position of every bounce captured by the Hawk-Eye system during that point, the serving player, the ongoing score, and the point winner. Altogether, this data set includes information on 1.8 million bounces from more than 1,800 men's singles matches, spanning various prestigious tournaments such as Grand Slam tournaments, Masters 1000 tournaments, and International series tournaments. This data set has been used in the past to study important economic questions such as risk management (\cite{Gauriot2017}), tournament incentives (\cite{Gauriot2019}), and mixed strategy play (\cite{Gauriot2023}). These papers have focused on the players' perspective, and the Hawk-Eye Base data set is well-suited for this purpose. However, this data set alone is insufficient to analyze umpires' performance, as it does not permit us to identify whether the final call comes from the umpire or follows a successful challenge.

\subsubsection{Challenge Data Set}

The Challenge data set was originally obtained from the ATP Tour and encompasses all challenges recorded during 35 tournaments across the initial three years following the introduction of Hawk-Eye (2006-2008). It captures comprehensive information on 2,784 challenges, including details about the players involved, the match itself, the specific point in the match when the challenge was made, and the outcome of each challenge. Of the 35 tournaments, we use the 28 that also appear in our Hawk-Eye Base data set. In order to compile this data set, \textcite{Abramitzky2012} undertook the arduous effort of collecting and compiling umpire match sheets. \textcite{Abramitzky2012} acknowledge the selection problem involved in only observing the challenged points (around 2.6\% of total points). Nevertheless, they are able to draw inferences on the optimality of decision-making from the players' standpoint by relying on the idea that a higher propensity to challenge implies challenging less conservatively. However, without the point-by-point data, it is hard to assess the umpire's performance once Hawk-Eye review was active. Combining the first two data sets allows us to analyze initial calls beyond the selected set of challenged points.

\subsubsection{Video Auditing}

In order to assess the quality of our merging algorithm, we audited full-match video replays using TennisTV, the official ATP streaming service. We identified 43 matches that appeared in both the Hawk-Eye Base and Challenge data sets, providing us with an assessment of how well a given merging algorithm would work on the 144 challenges witnessed during these matches.

Using variables such as match, point, distance, and who hit the ball, we developed a merging algorithm with a 99.3\% merge rate, for which only 4.9\% of the video-audited challenges were merged to an incorrect point. The algorithm consists of eight iterations, starting with the most stringent merging rule and gradually relaxing criteria for challenges that persist without merging.\footnote{A detailed explanation of the merging algorithm is provided in Appendix \ref{AppendixA2}.}

As an additional benefit, auditing match replays enabled us to test the effectiveness of the four criteria we implemented to identify incorrect calls (two criteria for each type of mistake). We deem an in call to be incorrect when a player's stroke is recorded as bouncing out in the Hawk-Eye Base data set, yet they still win the point, or if the data set records at least three more strokes after an out bounce. We deem an out call to be incorrect if all the strokes of a point are recorded as in, and the player delivering the final hit loses, or if there is a second serve after the first one was recorded as in the Hawk-Eye Base data set.\footnote{Appendix \ref{AppendixA3} documents these criteria further.}

\section{Empirical Results}\label{sec:results}

We examine how umpiring decisions changed after Hawk-Eye review was introduced. We begin with raw evidence of a shift in the decision rule and the composition of errors. We then estimate the pattern across distances with call, point, and match controls and discuss identification and supporting diagnostics. Next, we present suggestive evidence within matches about possible mechanisms and consider the average mistake rate as a secondary outcome. Section \ref{sec:hetero} examines heterogeneity, and Section \ref{sec:review_returns} connects these findings to the evaluation of review systems.

\subsection{Error Reallocation Near the Line}

\begin{figure}[!htbp]
\centering
\includegraphics[width=5.5in]{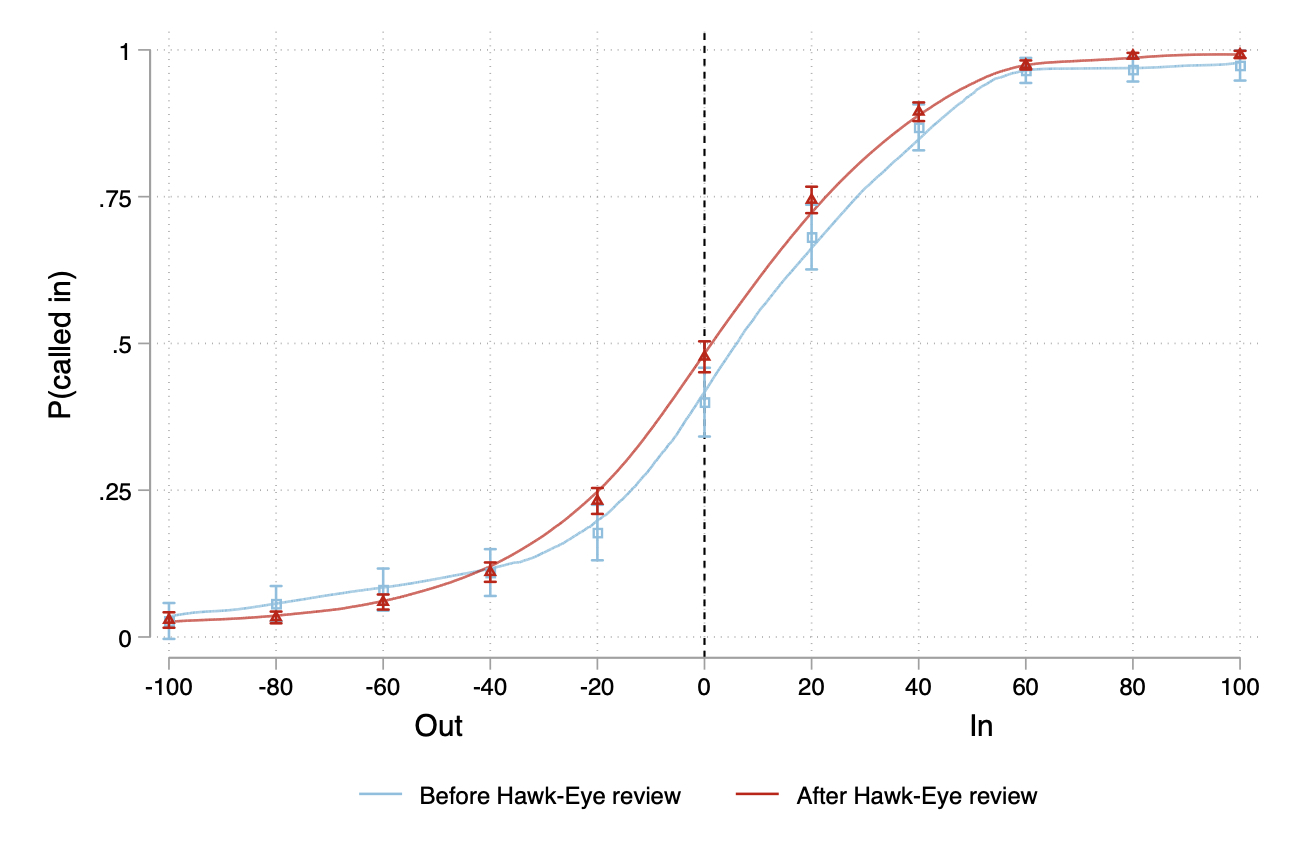}
\caption{The figure plots the probability that a ball is initially called ``in'' against its signed distance from the line. Negative distances indicate balls landing out and positive distances balls landing in. Each marker reports the mean call-in rate for observations within 10 mm of the indicated distance. For example, the marker at zero includes balls landing between -10 and 10 mm from the line. Bars denote 95\% confidence intervals. Curves are locally weighted scatterplot smoothing (LOWESS) fits, estimated separately before and after the introduction of Hawk-Eye review. The vertical dashed line denotes the court boundary.}
\label{fig:Mistakes}
\end{figure}

Figure \ref{fig:Mistakes} shows how umpire decisions differ with the introduction of Hawk-Eye review. It plots the probability that a ball is initially called in against its signed distance from the line, separately before and after review. As expected, the probability of an in call rises sharply as the true landing location moves from outside to inside the line. Near the boundary, the curve under review lies above the curve without review. Balls landing at similar locations were more likely to be initially called in under review.

Calling more marginal balls in increases mistakes for balls that land out and reduces mistakes for balls that land in. We therefore examine whether these opposing changes persist after accounting for observable differences across periods. We estimate the following specification:

\begin{equation}\label{eqn:EmpiricalSpecification}
\mathds{1}(Incorrect\, Call)_{ipm}= \alpha_0 + \sum_{b\in\mathcal{B}}\alpha_b
  \bm{(PostHK_m\times D_{ib})}  + \beta X_{ipm} + \gamma Y_m + \epsilon_{ipm}
\end{equation}

$\mathds{1}(Incorrect\, Call)$ is an indicator equal to 1 if call $i$ in point $p$ of match $m$ was incorrect. $PostHK_m$ equals 1 if Hawk-Eye review was active in the tournament edition containing match $m$. $D_{ib}$ identifies signed-distance bin $b$. The ten coefficients $\alpha_b$ estimate the adjusted change in mistakes within each bin. $X_{ipm}$ contains call and point controls: 20 mm signed-distance-bin fixed effects, an indicator for ball speed above the sample median, indicators for the server's point score, game number, and set number, and a tie-break indicator.\footnote{A tie-break is a one-off game held to decide the winner of a set when two players are locked at 6-6.} Distance and speed vary across calls within a point. $Y_m$ comprises indicators for the match round and tournament tier.\footnote{Because the period without review has no ATP 500 tournament editions, we combine the ATP 250 and ATP 500 tiers, so the tier control distinguishes Masters 1000 tournaments from ATP 250 and ATP 500 tournaments.} We cluster standard errors by tournament edition, yielding 35 clusters. This permits arbitrary correlation across calls and matches within the same tournament edition. Clustering by match is a plausible alternative because each match has its own officiating crew. It yields qualitatively similar and more precise results.

\begin{figure}
\centering
\includegraphics[width=5.5in]{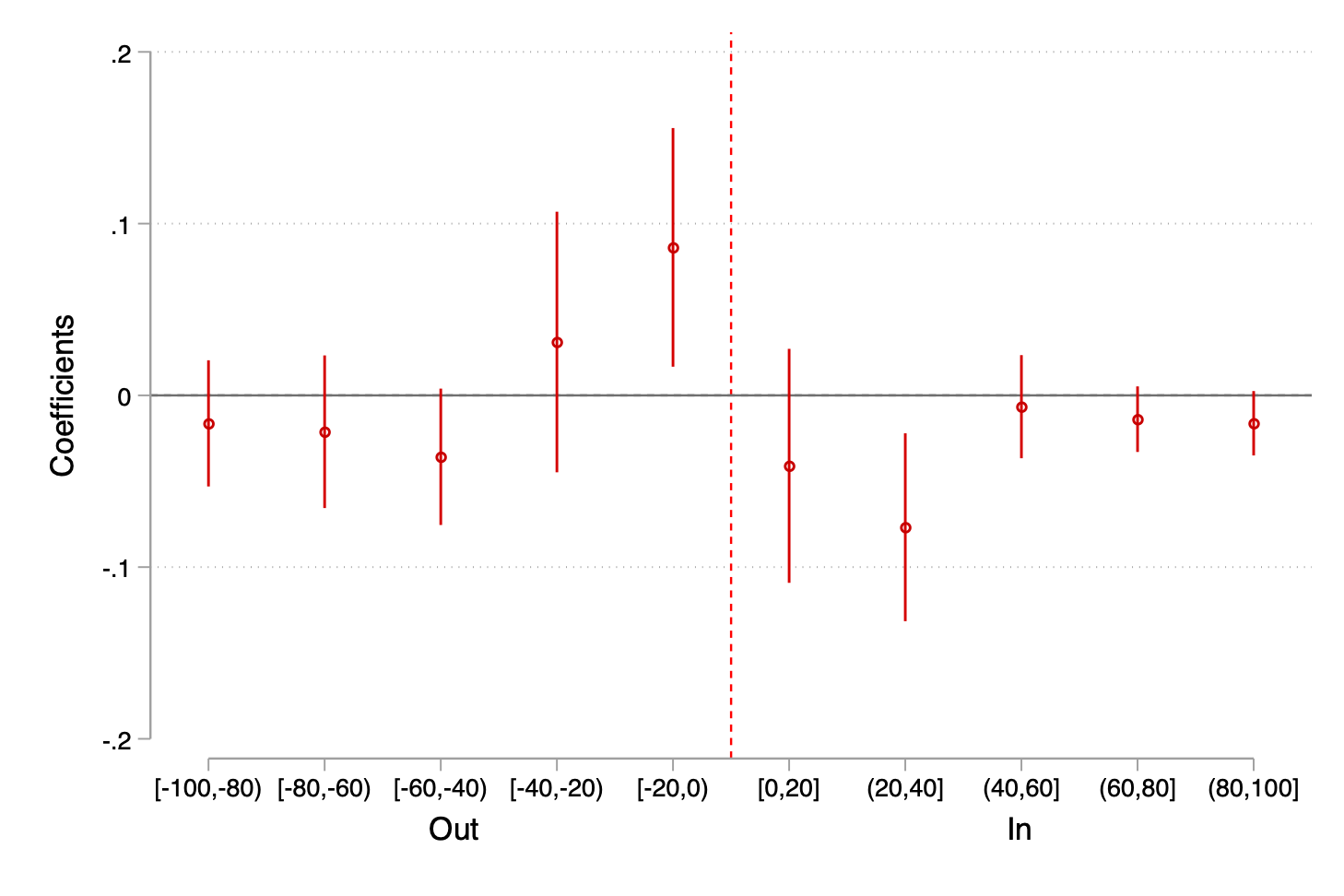}
\caption{Estimated change in the incorrect-call rate after the introduction of Hawk-Eye review, by proximity to the line. Each dot is the coefficient on the interaction between a distance-bin indicator and $PostHK$, which equals 1 when Hawk-Eye review is active. The regressions include call, point, and match controls. Standard errors are clustered at the tournament-edition level.}
\label{fig:HE}
\end{figure}

Figure \ref{fig:HE} reports the interaction coefficients for $PostHK_m$ and indicators $D_{ib}$ for each of the ten 20 mm distance bins, with 95\% confidence intervals based on standard errors clustered at the tournament-edition level. In the raw data, the mistake rate in the closest outside bin was 8.5 percentage points higher under review than the no-review baseline of approximately 25.0\%, a 34\% relative increase. After adjustment for call, point, and match characteristics, the estimated difference was 8.6 percentage points, while the mistake rate was lower for comparably close balls inside. Consistent with a shift in the decision threshold, the estimated change in the closest outside bin exceeds the average change in the first two inside bins by approximately 14.5 percentage points. This contrast is statistically significant using wild-cluster-bootstrap inference at the tournament-edition level ($p=0.034$).\footnote{Table \ref{tab:wildbootstrap} reports wild-cluster-bootstrap inference for the individual distance-bin estimates and cross-boundary contrasts.}

We examine the closest calls more directly by studying call-in rates for balls within 20 mm of the line, regardless of which side they landed on. The regression-adjusted call-in rate is 42.8\% without Hawk-Eye review and 49.0\% with review (see Figure \ref{fig:In}).\footnote{The fact that balls landing close to the line are more often called out is consistent with evidence from the visual-perception literature on moving objects (\cite{Whitney08}).} The difference also appears gradual. The fitted monthly trend is -0.62 percentage points before review and 0.61 percentage points afterward, and the estimated difference in slopes is 1.23 percentage points per month $(p=0.002)$.

Taken together, the raw and regression-adjusted estimates indicate a reallocation of initial errors near the boundary under Hawk-Eye review. Whether that reallocation improves system performance also depends on which errors are corrected and the burden of their remedies. We return to these outcomes in Section \ref{sec:review_returns}. We first state the identifying assumption underlying the empirical comparison and assess the evidence supporting it.

\subsection{Identification and Supporting Evidence}

Our estimates compare tournament editions with and without Hawk-Eye review. A causal interpretation requires that the conditional relationship between signed distance and the initial umpire call would not have changed systematically across these editions without review. Our central result on error reallocation requires a more specific assumption. No contemporaneous change in tournament composition, shot difficulty, or officiating would independently have shifted close calls toward ``in'' and thereby increased errors just outside the line while reducing errors just inside it. This assumption cannot be tested directly because the editions without review precede nearly all editions with review, tournament composition differs across regimes, and umpire identities are unavailable.

We nevertheless examine observable implications of the assumption and the sensitivity of the result to alternative specifications and samples. The distributions of signed distance and speed are similar across regimes (Figure \ref{fig:PlayerStrokes}). Controlling for speed, score, game, set, tie-break status, round, and tournament tier changes the estimated increase in errors for the closest outside bin only from 8.5 to 8.6 percentage points. The sign reversal around the line is also reproduced in separate distance-bin regressions and remains visible when the review period is divided into early and later subperiods (Figure \ref{fig:ES}).

Restricting the sample to named tournaments observed under both review regimes and including named-event fixed effects reduces the estimated cross-boundary contrast from 14.5 to 10.2 percentage points. The estimate remains economically sizable and of the same sign, although it is imprecisely estimated in the substantially smaller sample (wild-cluster-bootstrap $p=0.152$). No single named event drives the result. Estimates that omit each event in turn range from 7.7 to 12.5 percentage points (Table \ref{tab:sameevent}).

The cross-boundary contrast is also stable to richer controls for shot composition and difficulty. Table \ref{tab:composition} shows robustness to serve status, flexible speed controls, and player fixed effects. Following \textcite{Goldsmith-Pinkham}, we also estimate one-treatment-at-a-time regressions to address possible contamination bias (see Figure \ref{fig:Contamination}).

\begin{samepage}
The reported controls and sample restrictions preserve the central pattern. We next consider suggestive within-match evidence about how umpires adapted following public review of their decisions.

\end{samepage}

\FloatBarrier
\subsection{Within-Match Responses to Algorithmic Review Outcomes}
\label{sec:within_match}

The bundled review regime creates several possible mechanisms that our design cannot separate. Institutional guidance or supervisory pressure could have encouraged a new default, although the leading official we contacted confirmed that umpires were not instructed to make more in calls. Strategic adaptation to correction exposure is another possibility. Among initial mistakes on balls landing less than 100 mm from the line in editions with review, 33.8\% of mistaken in calls and 21.6\% of mistaken out calls were corrected (Table \ref{tab:review_error_paths}). These are correction fractions conditional on an initial mistake, not challenge rates among all calls. Hawk-Eye could also facilitate learning about call accuracy. Finally, immediate public reversal could create reputational concerns, shame or embarrassment, or a desire to avoid the more discretionary remedy that follows an incorrect out call. Broadcast coverage made errors visible before stadium review, but Hawk-Eye changed the immediacy, publicity, and consequences of correction at the same time.

To provide evidence more closely tied to adaptation after public correction, we examine whether prior review outcomes are associated with changes in subsequent calls within the same match. For balls landing within 20 mm of the line, we estimate whether the probability of an initial in call varies with the histories of four review outcomes: in-to-out reversals, out-to-in reversals, upheld in calls, and upheld out calls. An in-to-out reversal corrects an initial in call to out; an out-to-in reversal corrects an initial out call to in. Our baseline specification includes all four histories jointly, with separate indicators for one, two, and three or more prior events of each type, relative to zero prior events. The model includes call and point controls and match fixed effects, which absorb stable features of the officiating environment within each match. By using within-match variation, this analysis provides evidence complementary to the earlier comparisons across review regimes. Each coefficient captures an association conditional on the other feedback histories and these controls. This specification also allows us to compare adjustment following reversal with adjustment following review that upholds the initial call. Because challenges and their outcomes are endogenous, these comparisons provide suggestive evidence rather than a stand-alone causal test.

\begin{figure}[!htbp]
\centering

\begin{subfigure}[t]{0.82\textwidth}
    \centering
    \includegraphics[width=\textwidth]{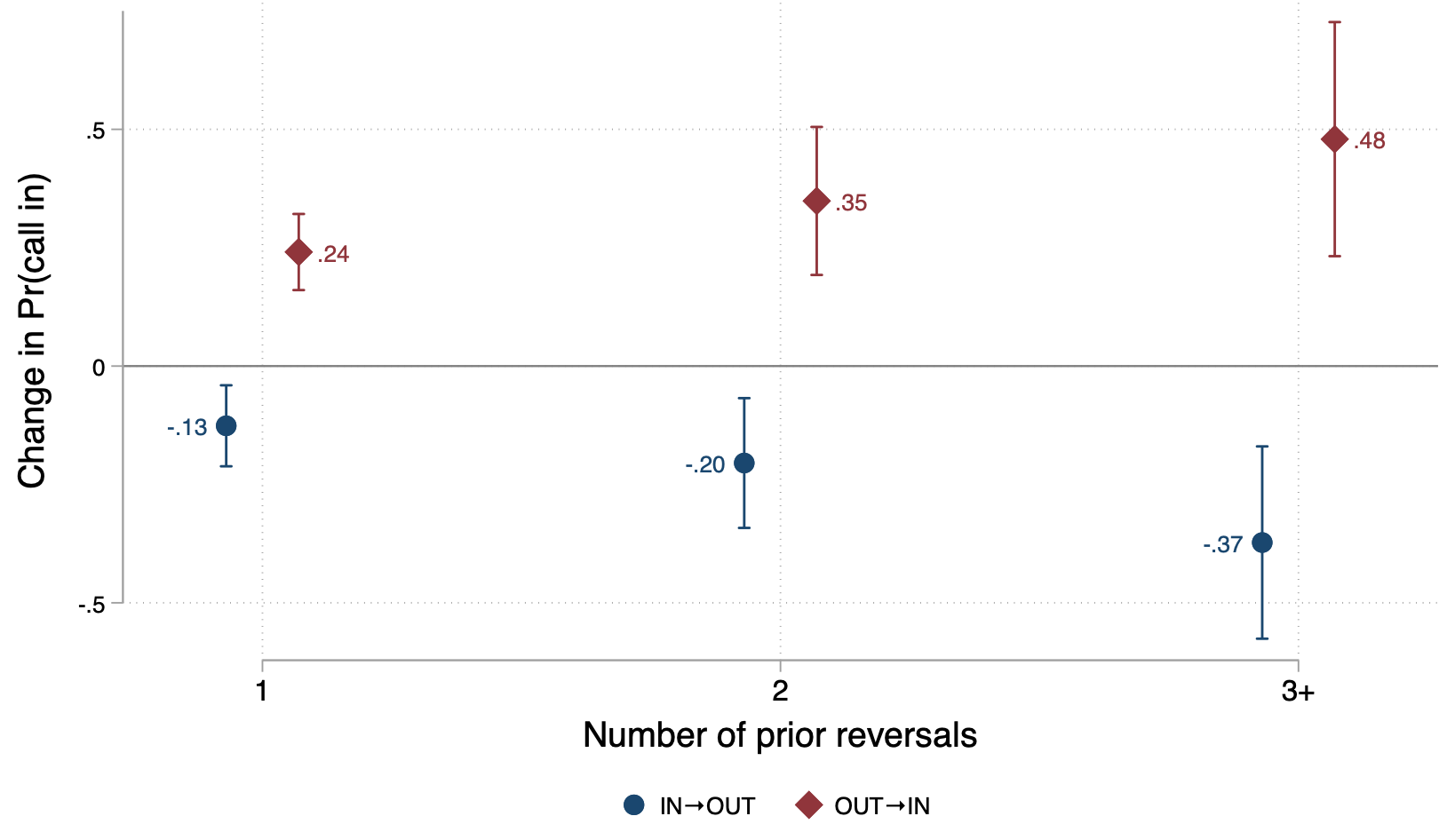}
    \caption{Successful challenges (reversals)}
    \label{fig:prior_reversals}
\end{subfigure}
\par\vspace{0.5em}
\begin{subfigure}[t]{0.82\textwidth}
    \centering
    \includegraphics[width=\textwidth]{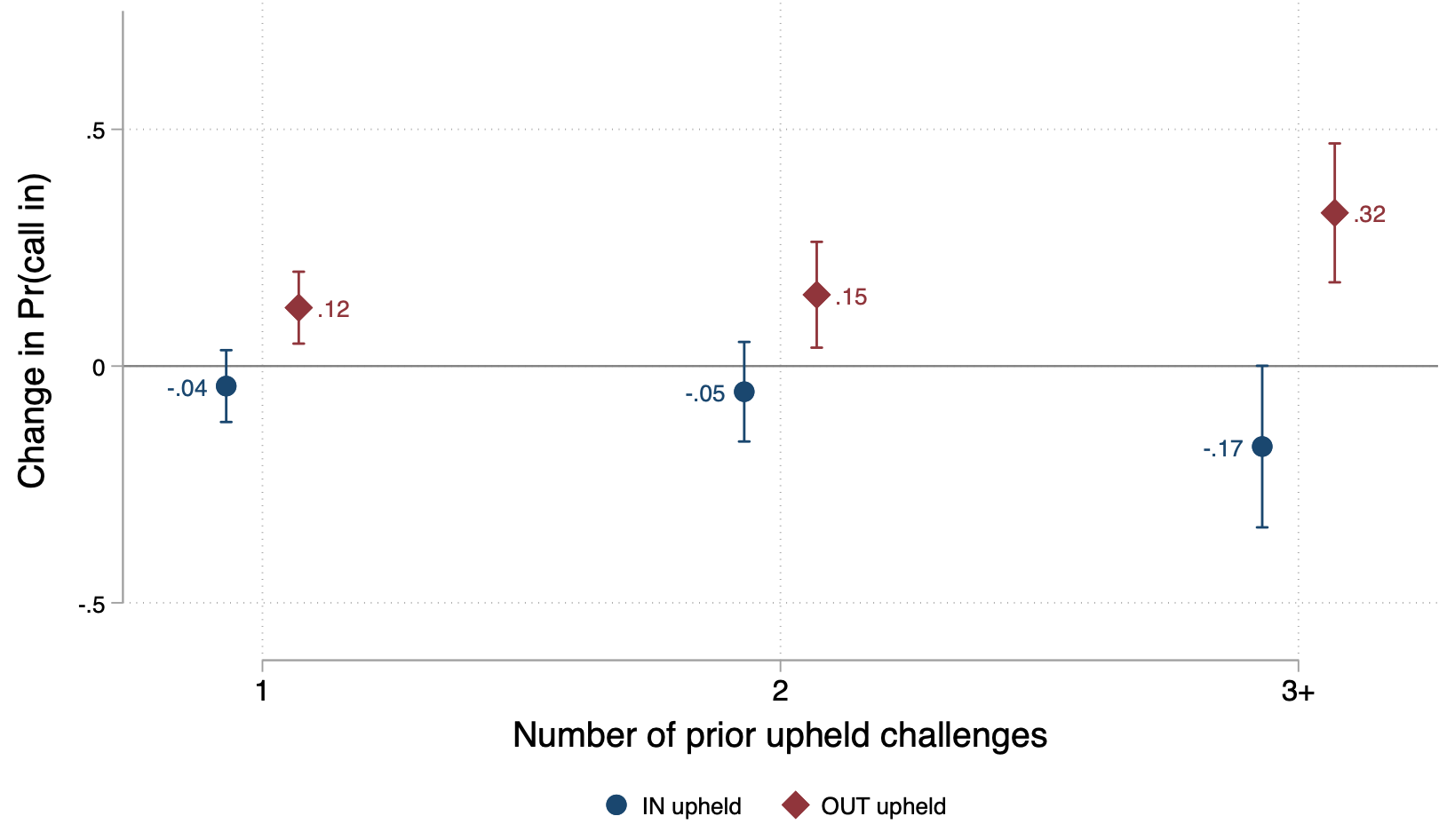}
    \caption{Unsuccessful challenges (upheld calls)}
    \label{fig:prior_upheld}
\end{subfigure}

\caption{
The figure reports coefficients from a linear probability model estimated on
post-Hawk-Eye calls for balls landing within 20 mm of the line. The dependent
variable equals one if the ball is initially called in. Panel (a) reports
responses to prior successful challenges that reversed the initial call. Panel
(b) reports responses to prior unsuccessful challenges that upheld the initial
call. Both panels display coefficients from a single regression that jointly includes indicators for one, two, and three or more prior events of each of the four feedback types. Each coefficient is relative to zero prior events of that type, conditional on the other feedback histories. Table \ref{tab:reversal_asymmetry}, Panel A, reports asymmetry tests calculated from this same regression. Counts are capped at three, with the final category pooling three or more events. The specification includes match fixed effects
and call and point controls. Standard errors are clustered at the match level.
Bars denote 95\% confidence intervals.}
\label{fig:hk_reversals_adjustment}
\end{figure}

First, we ask whether subsequent calls move after a reversal. Figure~\ref{fig:hk_reversals_adjustment}(a) compares calls following prior reversals with calls made before any reversal of the corresponding type within the match. Following in-to-out reversals, umpires become less likely to call later close balls in. Following out-to-in reversals, they become more likely to do so. In both directions, subsequent calls move toward avoiding the previously corrected error.

\begin{samepage}
Second, we ask whether the response is larger in one direction. We compare the increase in the probability of calling in after an out-to-in reversal with the absolute decline after an in-to-out reversal (Table~\ref{tab:reversal_asymmetry}). When the cumulative number of each reversal type enters linearly, the response to an additional out-to-in reversal is 8.9 percentage points larger in magnitude ($p=0.032$). Allowing separate responses after one, two, and three or more prior reversals gives differences of 11.5, 14.4, and 10.7 percentage points, respectively. These flexible estimates point in the same direction but are less precise.
\par
\end{samepage}

Third, we compare reversals with upheld challenges involving the same initial-call direction, which allows us to study whether a reversal is associated with additional adjustment beyond that accompanying review itself. Both successful and unsuccessful challenges subject the initial call to review, but only successful challenges reverse it. For each initial-call direction, we therefore compare the coefficient on prior reversals with the coefficient on prior challenges that upheld the call (Table~\ref{tab:reversal_upheld}). Relative to an upheld in call, an in-to-out reversal is associated with an additional 6.7 percentage point shift toward out ($p=0.078$). Relative to an upheld out call, an out-to-in reversal is associated with an additional 11.5 percentage point shift toward in ($p=0.004$). The two reversal-minus-upheld-call differences are jointly significant ($p=0.004$).

These patterns are consistent with adaptation to public correction and with a stronger response when the original error incorrectly stopped play. We next ask whether the reallocation was accompanied by a deterioration in average initial-call accuracy, as might occur if umpires treated review as a substitute for effort.

\subsection{Overall Initial-Call Accuracy} \label{sec:overall}

A common concern about review and correction systems is that decision makers may treat the technology as a safety net and free ride on its ability to correct their mistakes, reducing effort and becoming less accurate before review. We examine initial-call accuracy, which reflects both information acquisition and the decision rule. Effort is not directly observed. We estimate an analogous pooled specification in which the ten $PostHK_m\times D_{ib}$ interactions in equation~\eqref{eqn:EmpiricalSpecification} are replaced by a single $PostHK_m$ indicator.

For calls within 100 mm of the line, the uncontrolled estimate indicates a statistically significant 1.4 percentage point reduction in the initial mistake rate. Adding call, point, and match controls reduces the estimated effect size to 1.1 percentage points, about 8\% of the baseline mean, and it is no longer statistically significant at conventional levels (SE = 0.8 percentage points; Table~\ref{tab:regressions_pooled}). Mistake rates are lower in eight of the ten distance bins. To examine whether the average masks differences across tasks, Section~\ref{sec:serve} considers serves and non-serves separately.

\section{Examining Heterogeneity}\label{sec:hetero}

\subsection{Tournament Stage and Tier}

We examine descriptive heterogeneity across tournament stages and tiers, which differ in officiating assignments and stakes. These comparisons do not separately identify skill and stakes because stage and tier are imperfect proxies and umpire identities and skill are not directly observed. Umpires are rewarded for good performance with assignments to advanced-stage matches, as the pool of officials needed becomes smaller and organizers can be more selective as the tournament progresses. These selection dynamics within tournaments are likely to create variation in umpiring skill across tournament stages.

\begin{figure}
\centering
\includegraphics[width=5.5in]{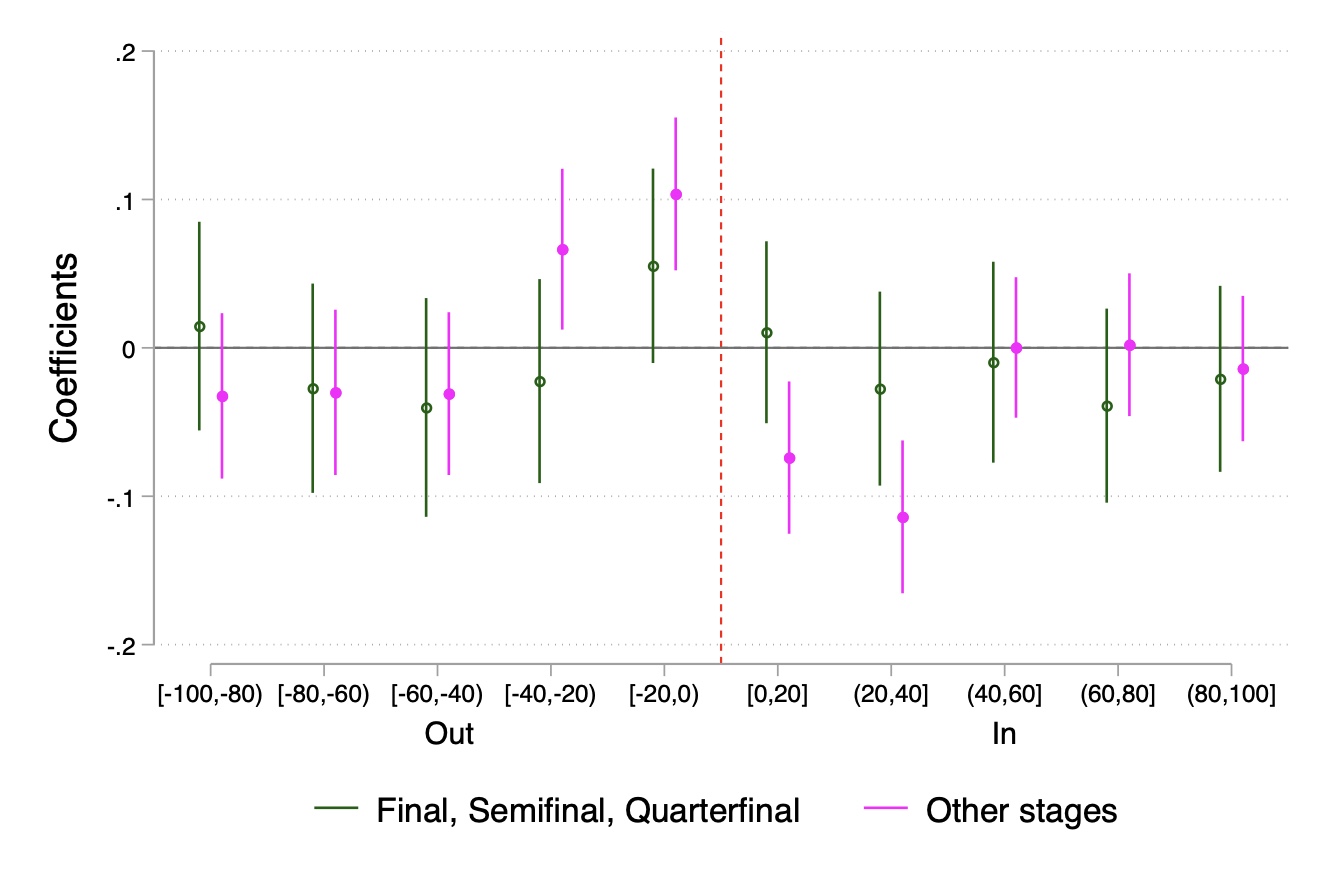}
\caption{Estimated change in the incorrect-call rate after the introduction of Hawk-Eye review, by proximity to the line and tournament stage. Matches at the final, semifinal, and quarterfinal stages are analyzed separately from earlier-stage matches. Each dot represents the coefficient on the interaction between a distance-bin indicator and $PostHK$.}
\label{fig:HE_FSQ}
\end{figure}

To examine the pattern by tournament stage, we split our sample into two groups. The first includes Final, Semifinal, and Quarterfinal matches where we expect assignments to favor more skilled umpires. The second contains the remaining matches. Figure \ref{fig:HE_FSQ} reports the estimates separately for these groups, rather than a formal test of differences between them. Separate significance tests do not establish a between-group difference. The estimated change in error composition near the line appears more pronounced for matches in earlier stages (magenta markers). The estimates for advanced-stage matches (green markers) are not statistically significant in any bin.

In a study on radiologists, \textcite{Chan2022} found that aversion to false negatives tends to be negatively related to radiologist skill. Specifically, lower-skilled radiologists are more likely to falsely diagnose a healthy patient with pneumonia (false positive) because this error is less costly than missing a diagnosis. This provides a possible interpretation of the tournament-stage pattern.

We also compare higher-tier tournaments (Masters 1000) and lower-tier tournaments (ATP 500 and 250). The official we consulted reported substantial overlap in officials across tiers, with assignments depending on geography, budget, availability, and logistics. Figure \ref{fig:HE_Cat} does not show a comparable pattern when contrasting higher- and lower-tier tournaments.

\subsection{Serves and Non-Serves}\label{sec:serve}

There is value in distinguishing between serves and non-serves,\footnote{The serve starts off every point in tennis, with players alternating serving each game.} as we can think of them as distinct tasks for the umpires. For umpires, the primary differences between serves and subsequent hits are speed and anticipation. Serves may be more challenging because they are much faster. Their average speed in our sample is 150 km/h compared to 83 km/h for non-serves.\footnote{Figure \ref{fig:Speed} shows the speed distribution for both types.} Serves also allow anticipation because umpires know the ball is about to be served and will most likely bounce in the service box. This appears to matter when we compare the slowest quartile of serves (132 km/h) with the fastest quartile of non-serves (97 km/h). Error rates are lower for serves despite their greater speed. Before Hawk-Eye review, the corresponding incorrect-call rates are 11.6\% for serves and 15.6\% for non-serves within 100 mm of the line, and 25.7\% and 40\%, respectively, within 20 mm.
Anticipation and differences in shot placement offer possible explanations for the lower error rates on serves.
% Not sure we need this anymore:
% In our sample, the average incorrect rate during the period without Hawk-Eye review is similar for serves and non-serves: 13.9\% versus 13.8\% within 100 mm of the line and 32.5\% versus 33.3\% within 20 mm of the line, respectively. 

Figure~\ref{fig:Mistakes_byserve} shows changes in error composition for both serves and non-serves. For serves, the main pattern is a reallocation of errors. For non-serves, it is a broader decline in mistakes.

\begin{table}%[htbp]
\centering
\def\sym#1{\ifmmode^{#1}\else\(^{#1}\)\fi}
\small
\scalebox{.9}{
\begin{tabular}{l*{8}{c}}
\hline\hline
                &\multicolumn{1}{c}{(1)}&\multicolumn{1}{c}{(2)}&\multicolumn{1}{c}{(3)}  &\multicolumn{1}{c}{(4)} &\multicolumn{1}{c}{(5)}&\multicolumn{1}{c}{(6)}&\multicolumn{1}{c}{(7)}&\multicolumn{1}{c}{(8)}\\
                &\multicolumn{4}{c}{Incorrect call (Serves)} &\multicolumn{4}{c}{Incorrect call (Non-serves)}\\
\hline
PostHK &  -0.006 &  -0.002 & 0.000 & 0.000 & -0.024\sym{**} & -0.021\sym{**} & -0.023\sym{**}  & -0.023\sym{*} \\
                & (0.010)   & (0.009)   & (0.009)   & (0.009) & (0.010)  & (0.010)  & (0.010) & (0.012)         \\
[1em]
\shortstack[l]{Call and point\\controls} & & X & X & X & & X & X & X \\
[.5em]
Match controls       &    &    &   X & X &  &    &   X &   X  \\
[.5em]
% DAVID VERIFY: Identify the SE estimator (conventional OLS or heteroskedasticity-robust) and test reference distribution in columns 1--3 and 5--7; blank cluster entries do not establish the method.
Cluster level & & & & \shortstack{Tournament\\edition} & & & & \shortstack{Tournament\\edition} \\
\hline 
N              & 9,253  &  9,253  &  8,990  &  8,990  &  7,559  &  7,559  &  7,345  &  7,345  \\
[.5em]
Baseline mean           &  .139  &  .139   &  .137  &  .137  &  .138  &  .138   &  .136  &  .136  \\
\hline\hline
\multicolumn{9}{l}{\footnotesize Standard errors in parentheses}\\
\multicolumn{9}{l}{\footnotesize \sym{*} \(p<.10\), \sym{**} \(p<.05\), \sym{***} \(p<.01\)}\\
\end{tabular}
}
\caption{OLS regressions of umpire mistakes for balls bouncing within 100 mm of the line separately for serves and non-serves. $PostHK=1$ if Hawk-Eye review is active. Call and point controls include 20 mm signed-distance-bin fixed effects, an indicator for ball speed above the sample median, indicators for the server's point score, game number, and set number, and a tie-break indicator. Match controls include round and tournament tier. Specifications with match controls exclude 263 serves and 214 non-serves relative to those without match controls. Coefficients, standard errors, and baseline means are in probability units (0.01 is one percentage point).}
\label{tab:regressions100sns}
\end{table}

Table \ref{tab:regressions100sns} reports the pooled specification described in Section \ref{sec:overall}, estimated separately for serves and non-serves within 100 mm of the line. Although separate significance levels do not establish a difference between tasks, the estimates suggest different patterns for serves and non-serves. We do not find evidence of a change in average performance for serves. For non-serves, the estimated mistake rate declines by 2.3 percentage points, a 17\% reduction relative to the baseline mean of 0.136. With tournament-edition clustering, its standard error is 1.2 percentage points.

If we restrict attention to balls landing between 20 and 100 mm from the line, Table \ref{tab:regressions10020} shows statistically significant reductions in mistakes for both serves and non-serves. The closest calls look different. Table \ref{tab:regressions20} restricts the sample to bounces within 20 mm, where perceptual difficulty is greatest. For serves in this range, the estimated mistake rate increases by 7.3 percentage points, or 22.9\% relative to baseline. For these serves, review is associated with worse initial-call accuracy.

\begin{table}[!htbp]
\begin{subtable}{0.95\textwidth}
\centering
\def\sym#1{\ifmmode^{#1}\else\(^{#1}\)\fi}
\small
\resizebox{\linewidth}{!}{
\begin{tabular}{l*{8}{c}}
\hline\hline
                &\multicolumn{1}{c}{(1)}&\multicolumn{1}{c}{(2)}&\multicolumn{1}{c}{(3)}  &\multicolumn{1}{c}{(4)} &\multicolumn{1}{c}{(5)}&\multicolumn{1}{c}{(6)}&\multicolumn{1}{c}{(7)}&\multicolumn{1}{c}{(8)}\\
                &\multicolumn{4}{c}{Incorrect call (Serves)} &\multicolumn{4}{c}{Incorrect call (Non-serves)}\\
\hline
PostHK &  -0.020\sym{**} &  -0.018\sym{**}  & -0.017\sym{**}  & -0.017\sym{**} & -0.012\sym{} & -0.017\sym{*} & -0.020\sym{**}  & -0.020\sym{**} \\
                & (0.008)   & (0.008)   & (0.008)   & (0.009) & (0.009)  & (0.009)  & (0.009) & (0.009)         \\
[1em]
\shortstack[l]{Call and point\\controls} & & X & X & X & & X & X & X \\
[.5em]
Match controls       &    &    &   X & X &  &    &   X &   X  \\
[.5em]
% DAVID VERIFY: Identify the SE estimator and test reference distribution in columns 1--3 and 5--7, as in Table 2.
Cluster level & & & & \shortstack{Tournament\\edition} & & & & \shortstack{Tournament\\edition} \\
\hline 
N              & 7,449  &  7,449  &  7,238  &  7,238  &  6,047  &  6,047  &  5,875  &  5,875  \\
[.5em]
Baseline mean           &  .092  &  .092   &  .091   &  .091  &  .082  &  .082   &  .080  &  .080  \\
\hline\hline
\multicolumn{9}{l}{\footnotesize Standard errors in parentheses}\\
\multicolumn{9}{l}{\footnotesize \sym{*} \(p<.10\), \sym{**} \(p<.05\), \sym{***} \(p<.01\)}\\
\end{tabular}
}
\par\smallskip
\caption{For balls bouncing 20-100 mm away from the line.}
\label{tab:regressions10020}
\end{subtable}
\par\medskip
\begin{subtable}{0.95\textwidth}
\centering
\def\sym#1{\ifmmode^{#1}\else\(^{#1}\)\fi}
\small
\resizebox{\linewidth}{!}{
\begin{tabular}{l*{8}{c}}
\hline\hline
                &\multicolumn{1}{c}{(1)}&\multicolumn{1}{c}{(2)}&\multicolumn{1}{c}{(3)}  &\multicolumn{1}{c}{(4)} &\multicolumn{1}{c}{(5)}&\multicolumn{1}{c}{(6)}&\multicolumn{1}{c}{(7)}&\multicolumn{1}{c}{(8)}\\
                &\multicolumn{4}{c}{Incorrect call (Serves)} &\multicolumn{4}{c}{Incorrect call (Non-serves)}\\
\hline
PostHK &  0.063\sym{**} &  0.063\sym{**}  & 0.073\sym{**}  & 0.073\sym{***} & -0.035\sym{} & -0.037\sym{} & -0.031\sym{}  & -0.031\sym{} \\
                & (0.031)   & (0.031)   & (0.032)   & (0.021) & (0.031)  & (0.031)  & (0.033) & (0.032)         \\
[1em]
\shortstack[l]{Call and point\\controls} & & X & X & X & & X & X & X \\
[.5em]
Match controls       &    &    &   X & X &  &    &   X &   X  \\
[.5em]
% DAVID VERIFY: Identify the SE estimator and test reference distribution in columns 1--3 and 5--7, as in Table 2.
Cluster level & & & & \shortstack{Tournament\\edition} & & & & \shortstack{Tournament\\edition} \\
\hline 
N              & 1,804  &  1,804  &  1,752  &  1,752  &  1,512  &  1,512  &  1,470  &  1,470  \\
[.5em]
Baseline mean           &  .325  &  .325   &  .319   &  .319  &  .333  &  .333   &  .325  &  .325  \\
\hline\hline
\multicolumn{9}{l}{\footnotesize Standard errors in parentheses}\\
\multicolumn{9}{l}{\footnotesize \sym{*} \(p<.10\), \sym{**} \(p<.05\), \sym{***} \(p<.01\)}\\
\end{tabular}
}
\par\smallskip
\caption{For balls bouncing within 20 mm of the line.}
\label{tab:regressions20}
\end{subtable}
\caption{OLS regressions of umpire mistakes, separately for serves and non-serves. $PostHK=1$ if Hawk-Eye review is active. Call and point controls include 20 mm signed-distance-bin fixed effects, an indicator for ball speed above the sample median, indicators for the server's point score, game number, and set number, and a tie-break indicator. Match controls include round and tournament tier. Specifications with match controls exclude 211 serves and 172 non-serves in panel (a), and 52 serves and 42 non-serves in panel (b), relative to those without match controls. Coefficients, standard errors, and baseline means are in probability units (0.01 is one percentage point).}
\end{table}

%\FloatBarrier
\section{Organizational Returns to Technology Investment}
\label{sec:review_returns}

Organizations considering a new technology need to assess the productivity
gains it can deliver relative to the cost of adoption and operation.
For a review system, one approach is to apply its correction capability
to a fixed set of human decisions and calculate the resulting reduction
in mistakes. This calculation can misvalue the investment if workers
change their decisions when the system is introduced. Their response may
improve initial accuracy, reduce care, or shift mistakes toward error
types that matter more or less to the organization.

We compare the returns to review when human decisions are held fixed
with the returns after allowing initial error rates to change.
Both calculations use the same correction fractions and unit costs.
We value performance through the losses from remaining errors and the
costs of remedying corrected mistakes. The comparison makes explicit
how the human response at implementation can strengthen or weaken the
case for adopting the technology.

\subsection{Two Investment Calculations}
\label{subsec:review_two_calculations}

For each error type $j\in\{\mathrm{in},\mathrm{out}\}$, let $E_j^0$ be
the share of all calls initially wrong without review and $E_j^1$ the
share initially wrong with review. A mistaken \enquote{in} call
classifies an out ball as in; a mistaken \enquote{out} call classifies
an in ball as out.

The human decides first. Review then corrects a fraction
$\eta_j\in[0,1]$ of the initial mistakes of type $j$. The share of all
calls corrected is therefore $C_j=\eta_jE_j^1$, and the share left
incorrect is $(1-\eta_j)E_j^1$. In tennis, $\eta_j$ includes whether an initial
mistake is challenged and overturned. It measures correction by the
review process, rather than Hawk-Eye's technical accuracy alone.
As in the empirical analysis, we treat Hawk-Eye as the accuracy
benchmark and assume review introduces no new line-classification
errors.

Let $w_j$ be the loss from an error left uncorrected and $k_j$ the
cost of remedying a corrected error. These are organizational costs,
not officials' private costs of making or reversing a mistake. For a
social evaluation, the weights can also include broader harms. All costs and
losses are expressed in comparable monetary units. The expected cost
per initial mistake after review, excluding the cost of providing
review, is
\begin{equation}
    \ell_j=(1-\eta_j)w_j+\eta_jk_j.
    \label{eq:review_expected_loss}
\end{equation}
Correcting a decision need not eliminate all its consequences.
For example, reversing a mistaken \enquote{out} call restores the
line classification but may still require a replay or another ruling
about who receives the point. The term $k_j$ allows such consequences
to remain even after the classification is corrected.

Let $H$ be the additional cost per call of providing and conducting
review, including reviews that uphold the initial call but excluding
remedy costs already counted in $k_j$. Without review, expected losses
are $\sum_j E_j^0w_j$ per call.

The engineering benchmark applies review to the original human error
rates, giving a cost of $H+\sum_jE_j^0\ell_j$. The calculation including
human responses instead applies review to $E_j^1$, giving a cost of
$H+\sum_jE_j^1\ell_j$. The savings available to cover $H$ are
\begin{equation}
\begin{aligned}
    B_{\mathrm{eng}}
        &=\sum_jE_j^0(w_j-\ell_j)
          =\sum_j\eta_jE_j^0(w_j-k_j),\\
    B_{\mathrm{resp}}
        &=\sum_j\left(E_j^0w_j-E_j^1\ell_j\right).
\end{aligned}
\label{eq:review_investment_values}
\end{equation}
Review lowers costs when $H<B_{\mathrm{eng}}$ under the engineering
calculation and when $H<B_{\mathrm{resp}}$ under the calculation
including human responses. If the relevant $B$ is nonpositive, these
savings cannot justify a positive review expenditure. Over $N$ calls,
a positive $B$ supports expenditure of up to $NB$ before net savings
are exhausted.

The difference between the two calculations is
\begin{equation}
    B_{\mathrm{resp}}-B_{\mathrm{eng}}
        =\sum_j(E_j^0-E_j^1)\ell_j.
    \label{eq:review_investment_gap}
\end{equation}
Holding error composition fixed, fewer initial mistakes increase the
amount an organization can justify spending on review. A shift between error types can raise or lower
that amount, depending on their expected costs after correction.

Both calculations hold $\eta_j$, $w_j$, and $k_j$ fixed to isolate
the role of initial error rates. The engineering benchmark therefore
predicts that a share $\eta_jE_j^0$ of all calls is corrected for each
error type, while the calculation including human responses predicts
a share $\eta_jE_j^1$. The share of calls corrected changes with the
initial error rate, even when the fraction of mistakes corrected is
unchanged.

\subsection{Calibrating the Investment Comparison}
\label{subsec:review_calibration}

Table~\ref{tab:review_error_paths} gives the inputs for calls landing
within 100~mm of the line. We estimate each correction fraction from
corrected errors divided by initial errors in the review regime and
use that fraction in both calculations. These are unadjusted
descriptive shares, rather than the controlled estimates reported
earlier. Appendix~\ref{app:structural_moments} reports the underlying
call and correction counts.

\begin{table}[tbp]
\centering
\small
\begin{tabular*}{\linewidth}{@{\extracolsep{\fill}}lrr@{}}
\toprule
Input & \shortstack{Mistaken\\in} & \shortstack{Mistaken\\out}\\
\midrule
Initial errors without review, $E_j^0$ (\%)
    & 5.61 & 8.28\\
Initial errors with review, $E_j^1$ (\%)
    & 6.19 & 6.27\\
Corrected errors with review, $C_j$ (\%)
    & 2.09 & 1.35\\
Fraction of initial errors corrected, $\eta_j=C_j/E_j^1$ (\%)
    & 33.8 & 21.6\\
\bottomrule
\end{tabular*}
\caption{Inputs to the two investment calculations for balls landing
within 100~mm of the line. The first three rows are shares of all
calls in the corresponding regime, expressed in percent. The last
row reports the percentage of initial mistakes corrected, calculated
as $\eta_j=C_j/E_j^1$. The sample contains 2,622 calls without review
and 14,190 with review, with equal weight per call and calls exactly
100~mm from the line excluded. Correction fractions use unrounded
shares and may therefore differ slightly from ratios calculated
using the displayed values.}
\label{tab:review_error_paths}
\end{table}

Per 100 calls, initial mistaken in calls rise from 5.61 without review
to 6.19 with review, while mistaken out calls fall from 8.28 to 6.27.
Review corrects 2.09 mistaken in calls and 1.35 mistaken out calls,
leaving approximately 4.10 and 4.92, respectively. Thus, the change in
initial decisions and the subsequent correction process both contribute
to the mix of remaining errors.

To assess the role of human responses, the engineering benchmark applies
the same correction fractions to initial error rates without review.
It predicts 3.68 corrections per 100 calls, leaving 10.20 errors.
Using the observed initial decisions instead gives 3.45 corrections
and 9.02 remaining errors. Thus, accounting for the observed human
decisions implies fewer corrections but a larger reduction in remaining
mistakes.

Applying these inputs to equation~\eqref{eq:review_investment_values},
the savings per 100 calls are approximately
\begin{equation}
\begin{aligned}
    100B_{\mathrm{eng}}
        &\approx 1.89(w_{\mathrm{in}}-k_{\mathrm{in}})
                 +1.79(w_{\mathrm{out}}-k_{\mathrm{out}}),\\
    100B_{\mathrm{resp}}
        &\approx 1.50w_{\mathrm{in}}+3.36w_{\mathrm{out}}
                 -2.09k_{\mathrm{in}}-1.35k_{\mathrm{out}}.
\end{aligned}
\label{eq:review_calibrated_values}
\end{equation}
The increase in initial mistaken \enquote{in} calls reduces the
return relative to the engineering benchmark. Fewer initial mistaken
\enquote{out} calls increase it. Which effect dominates depends on
the costs attached to each type. Adoption depends on their combined
contribution, rather than requiring each category to generate
savings separately.

\subsection{Investment Thresholds Using Hawk-Eye Costs}
\label{subsec:review_investment_thresholds}

We use reported Hawk-Eye charges from the study period to illustrate
the investment implications. \textcite{Fischetti2007} reports that
setup and operation cost \$40,000--\$50,000 per court-week.
For simplicity, we use the upper end of this range and set
$K=\$50{,}000$ throughout the calculations below.
Assuming 25 matches per court-week and 25 calls within 100~mm of
the line per match gives $N=625$ calls. Considering the system
charge alone therefore sets $H=K/N=\$80$ per call.

Table~\ref{tab:review_investment_thresholds} compares three cases
under the same charge and usage. In each case, let
$w=w_{\mathrm{out}}$. The first two cases assign equal losses to
both error types. The third assigns a larger loss to mistaken
\enquote{in} calls. All monetary values in this illustration are
in 2007 dollars.

\begin{table}[tbp]
\centering
\small
\begin{tabular*}{\linewidth}{@{\extracolsep{\fill}}p{0.52\linewidth}rr@{}}
\toprule
Valuation assumption
& \shortstack{Engineering\\benchmark}
& \shortstack{Including human\\responses}\\
\midrule
Equal losses; costless remedies:\newline
$w_{\mathrm{in}}=w_{\mathrm{out}}=w$, $k_j=0$
    & \$2,174 & \$1,645\\
\addlinespace
Equal losses; remedy cost equals error loss:\newline
$w_{\mathrm{in}}=w_{\mathrm{out}}=w$, $k_j=w$
    & --- & \$5,650\\
\addlinespace
Unequal losses; costless remedies:\newline
$w_{\mathrm{in}}=10w$, $w_{\mathrm{out}}=w$, $k_j=0$
    & \$386 & \$435\\
\bottomrule
\end{tabular*}
\caption{Minimum loss per mistaken \enquote{out} call needed to cover
a \$50,000 court-week charge for Hawk-Eye, in 2007 dollars. All rows
use $N=625$ calls and the correction fractions in
Table~\ref{tab:review_error_paths}. A dash indicates that no finite
$w$ covers a positive charge under that calculation. The 10:1 loss
ratio is illustrative. Thresholds use unrounded shares and cover
the assumed setup and operation charge alone; additional costs
raise them, while other benefits lower them.}
\label{tab:review_investment_thresholds}
\end{table}

\paragraph{Equal losses and costless remedies.}
When both error types carry the same loss and remedies are costless,
the engineering benchmark values 3.68 fewer errors per 100 calls,
compared with 4.86 when human responses are included.
The latter therefore supports about 32\% more expenditure for a
given $w$ and $N$. At the assumed \$50,000 charge, the required
loss per error falls from \$2,174 to \$1,645.

\paragraph{Equal losses and costly remedies.}
The case $k_j=w_j=w$ provides a second benchmark. Each corrected
mistake removes loss $w$ but incurs an equal remedy cost, so
correction alone yields no net saving. The two calculations reduce to
\begin{equation}
    B_{\mathrm{eng}}=0,
    \qquad
    B_{\mathrm{resp}}
        =w\sum_j(E_j^0-E_j^1)
        \approx 0.014160w.
    \label{eq:review_equal_remedy_costs}
\end{equation}
The engineering benchmark cannot justify a positive review
expenditure. Including human responses can do so because the
descriptive inputs contain fewer initial mistakes. Covering the
assumed charge requires a loss of approximately \$5,650
per uncorrected error.

\paragraph{Unequal losses across error types.}
Equal loss weights need not be appropriate in other applications.
\textcite{Mallett2012} emphasize that false-positive and false-negative
diagnoses can have different clinical consequences, and that their
relative costs depend on the clinical setting and whose perspective
is being evaluated. In criminal adjudication, convicting an innocent
defendant and acquitting a guilty defendant have different social
consequences. Harlan's concurrence in \emph{In re Winship} explicitly
relates standards of proof to these relative losses
(\cite[370--372]{Winship1970}). These examples motivate allowing
loss weights to differ.

To isolate the role of unequal losses, again set
$k_{\mathrm{in}}=k_{\mathrm{out}}=0$.
Equation~\eqref{eq:review_investment_gap} then gives approximately
\begin{equation}
    100(B_{\mathrm{resp}}-B_{\mathrm{eng}})
        \approx -0.39w_{\mathrm{in}}+1.57w_{\mathrm{out}}.
    \label{eq:review_asymmetric_losses}
\end{equation}
The engineering calculation overstates the value of review once
the loss from a mistaken \enquote{in} call exceeds roughly four
times the loss from a mistaken \enquote{out} call. The improvement
in the latter then fails to compensate for the greater weight
attached to the increase in the former.

For illustration, set $w_{\mathrm{in}}=10w$ and
$w_{\mathrm{out}}=w$. The savings per 100 calls are approximately
$100B_{\mathrm{eng}}=20.73w$ and
$100B_{\mathrm{resp}}=18.41w$. The required value of
$w_{\mathrm{out}}$ rises from \$386 under the engineering
calculation to \$435 when human responses are included.

For a court-week charge of \$50,000, an organization assigning
losses of \$400 to a mistaken \enquote{out} call and \$4,000 to a
mistaken \enquote{in} call would pass the engineering investment
criterion but fail the criterion including human responses.
Although fewer errors remain after review in the latter calculation,
their weighted loss is higher than the engineering benchmark
predicts. The two calculations therefore give opposite adoption
recommendations for the same technology and charge.

\FloatBarrier
\begingroup
\clubpenalty=10000
\widowpenalty=10000
\brokenpenalty=10000

\section{Conclusion}\label{sec:conclusion}

Algorithmic review can change the decisions it is intended to correct. Following the introduction of Hawk-Eye review, officials became more inclined to call the closest balls “in.” The main finding is a shift in the composition of errors, with no evidence of a decline in average initial-call accuracy.

The direction of this shift is consistent with officials seeking to avoid mistaken “out” calls, whose reversal can require an additional, potentially contentious judgment about whether to award the point or replay it. Within matches, subsequent close calls also shift toward avoiding the type of mistake just overturned. Public scrutiny may increase officials’ concerns about these consequences, while also discouraging reliance on review as a safety net. Appendix~\ref{sec:ev} discusses the scope for extending these findings to other settings.

These responses matter for the returns to review technology. An engineering calculation that applies correction to a fixed set of human decisions overlooks changes in both the number and composition of initial errors. Our investment illustrations show that incorporating those changes can reverse the adoption decision in either direction, holding correction fractions and unit costs fixed within each comparison. Which calculation supports greater expenditure depends on the losses attached to different errors and the costs of remedying them. How organizations deploy review technology is therefore an important design decision. The rules governing review and correction can shape workers’ initial choices and, through those responses, the system’s performance and returns.

\par
\endgroup

\pagebreak
\begingroup
\setlength{\emergencystretch}{3em}
\renewcommand{\baselinestretch}{1.1}\selectfont
\raggedright
\printbibliography
\endgroup

\pagebreak

\appendix
\counterwithin{figure}{section}
\counterwithin{table}{section}
\section{Data Management}

\subsection{Exclusion of Clay Tournaments}\label{AppendixA1}

During our study period, Hawk-Eye review was not used at clay-court tournaments. One concern was that changes in the clay surface during play could affect calibration.\footnote{https://www.perfect-tennis.com/why-there-is-no-hawk-eye-on-clay/} Another was that Hawk-Eye's assessment could disagree with the visible ball mark. As noted in The Guardian, Lars Graf, one of the ATP's most experienced officials, explained the decision not to use Hawk-Eye on clay: ``We decided not to use Hawk-Eye on clay because it might not agree with the mark the umpire is pointing at.''\footnote{https://www.theguardian.com/lifeandstyle/2009/jun/27/tennis-hawk-eye}

Because Hawk-Eye review was not adopted in clay tournaments during our sample period, they may appear as an ideal control group for comparing umpire reactions in tournaments with Hawk-Eye review versus those on clay. However, there are several factors that led us to exclude clay tournaments from our analysis. For one, we are unable to determine from our data set which calls were corrected after examining the clay marks. In addition, before the introduction of Hawk-Eye review, players were allowed to request the chair umpire to review the ball bounce marks on the surface, and changing the call based on that was permitted, so a form of review system was available.

\subsection{Merging Algorithm}\label{AppendixA2}

Our merging algorithm was designed to identify the highest number of the 144 challenges (derived from the 43 matches with video replays) within the Hawk-Eye Base data set, while keeping the number of false positives (challenges matched into the wrong point) as low as possible. We used standard variables, such as set, game, score, distance difference, player hitting the ball, and whether it was a tie-break, in order to minimize overfitting. 

Our final merging algorithm comprises eight iterations, where we start using stricter criteria and gradually relax them in subsequent iterations. Iterations 5 and 6 target tie-break situations but produced no additional matches in the audited sample.

The algorithm merged 143 out of the 144 challenges, achieving a merge rate of $99.3\%$. Out of these, 136 were merged to the correct point, as validated through video auditing replays, resulting in an accuracy rate of $95.1\%$. We will now elaborate on the specific criteria used in each of the eight iterations, followed by Table \ref{tab:a1}, which summarizes the number of challenges successfully merged (and false positives) achieved in each iteration. 

\bigskip
\noindent
Merging criteria for each iteration:
\begin{itemize}
    \item []\textcolor{violet}{Iteration 1.} Same set, game, score, and player (with distance difference <35 mm).
    \item []\textcolor{violet}{Iteration 2.} Same set, game, and player (with distance difference <15 mm).
    \item []\textcolor{violet}{Iteration 3.} Same set, score, and player (with distance difference <15 mm).
    \item []\textcolor{violet}{Iteration 4.} Same game, score, and player (with distance difference <10 mm).
    \item []\textcolor{violet}{Iteration 5.} For tie-breaks: Same game and score (with distance difference <35 mm).
    \item []\textcolor{violet}{Iteration 6.} For tie-breaks: Same game. If multiple, pick closest in distance (with distance difference <15 mm).
    \item []\textcolor{violet}{Iteration 7.} Same set, game and player.  If multiple, pick closest point in score and then in distance (with distance difference <35 mm).
    \item []\textcolor{violet}{Iteration 8.} Same set and player.  If multiple, pick closest point in distance (with distance difference <35 mm).
\end{itemize}

\begin{table}
\centering
\begin{tabular}{c|ccc|c}
\hline\hline
Iteration & \shortstack{Correct\\matches} & \shortstack{False\\positives} & \shortstack{Correct matches among\\matched challenges (\%)} & \shortstack{Challenges\\left} \\ \hline
1  & 98 & 2 & 98 & 44 \\ 
2 & 25  & 2 & 93 & 17 \\ 
3  & 5  & 0 & 100 & 12 \\ 
4 & 1  & 0 & 100 & 11 \\ 
5  & 0  & 0 & --- & 11 \\ 
6  & 0  & 0 & --- & 11 \\ 
7  & 5  & 0 & 100 & 6 \\ 
8  & 2  & 3 & 40 & 1 \\ 
\hline\hline
\end{tabular}
\caption{Performance for each iteration of the merging algorithm. The percentage of correct matches divides correct matches by the sum of correct matches and false positives within each iteration, rounded to the nearest integer. A dash indicates that no challenges were matched in that iteration. Out of the 144 challenges audited, only one remained unmerged.}
\label{tab:a1}
\end{table}

\subsection{Identification of Incorrect Calls}\label{AppendixA3}

The Hawk-Eye Base dataset records the precise location of every ball bounce and identifies the winner of each point, but does not directly report the umpires' calls. We therefore reconstruct the original calls using four criteria. The first two identify points at which the umpire incorrectly called a ball in, while the final two identify points at which the umpire incorrectly called a ball out. During the period when players could challenge calls, we first recover the umpire's original call for each successful challenge and then assess whether any of the four criteria are met.

\noindent
\textbf{\textcolor{red}{Criterion 1.} A player hits the ball out but wins the point.}

We selected strokes where the first player to hit a ball out in a given point also won the point. %We identified 55 points that satisfy this criterion, and 54 of them were confirmed by the video replays to be mistakes, resulting in a $98\%$ accuracy rate.

\noindent
\textbf{\textcolor{red}{Criterion 2.} The point continued after an out.}

To complement Criterion 1, we also identify points in which a player hit a ball out, play continued, and that same player ultimately lost the point. Specifically, we examine instances in which at least three additional strokes were recorded after the ball landed out.\footnote{Video validation indicated that a threshold of three additional strokes is necessary. A threshold of two is too lenient and captures multiple cases in which the call was correct but the players continued hitting the ball after the point had ended.}%Out of the 25 points satisfying this criterion, 24 were umpire mistakes, yielding an accuracy rate of $96\%$.

%We also tested a more relaxed criterion that identifies points recording at least two more strokes after an out. However, this criterion proved to be too lenient, encompassing multiple cases where the point was correctly called, and players continued hitting the ball afterward. When the criterion is relaxed to require at least two more strokes, the accuracy rate drops to $50\%$. This adjustment results in the identification of one new mistake at the expense of 24 incorrectly identified instances. This is why we adhere to the 3+ rule.

\noindent
\textbf{\textcolor{red}{Criterion 3.} All the strokes of a point are in, and the player hitting last loses.}

We identify strokes where the last player to hit the ball in did not win the point. %Out of the 25 points identified in the auditing matches, 23 were confirmed to be mistakes, resulting in a $92\%$ accuracy rate.

\noindent
\textbf{\textcolor{red}{Criterion 4.} Observing a second serve after the first serve was in.}

%This criterion helps identify two types of instances where there is no winner in a point as a direct consequence of an umpire mistake.

This criterion identifies first serves that landed in but were incorrectly called out, leaving the point without a recorded winner and prompting a second serve.\footnote{For this criterion, we restrict the analysis to serves landing within 40 mm of the line because the criterion becomes less accurate beyond this threshold.} %Second, it detects points that lasted multiple strokes but had to be replayed because the umpire stopped the point by incorrectly calling a ball out. The accuracy rate for this criterion is $83.7\%$, with 36 of the 43 selected points identified as umpire mistakes. Some inaccurately selected points by this criterion include lets (serves that hit the top of the net and the point is replayed) and other unusual events (e.g., a server touching the line while serving, and the serve not counting). The occurrence of these events, and our ability to identify them, should not change with the introduction of Hawk-Eye.
 %For balls bouncing between 40-100 mm away from the line, this criterion has a $30\%$ accuracy rate, as only seven of the 23 points audited were actual umpire mistakes.

\pagebreak
\section{External Validity}\label{sec:ev}

To provide a more complete assessment of external validity, we build around the four transparency conditions given as an \textit{onus probandi} in \textcite{List2020}.

\subsection{Selection}

This condition concerns the representativeness of the studied group compared to the underlying population, in particular in ``observables that might impact preferences, beliefs, or individual constraints.'' The evidence concerns experienced officials making rapid perceptual judgments under public review, but does not establish how responses vary with expertise or professional identity.

\subsection{Attrition}

This condition asks whether motivational or incentive differences affect continued participation. A leading official at the time of Hawk-Eye's introduction confirmed that umpire training and assessment, the pool of umpires, and the positioning and instructions given to line judges remained stable during our study period. In other settings, introducing algorithmic review and public correction could cause selective attrition. Workers who dislike correction most may be especially likely to leave, so estimates among those who remain could understate the response in the original workforce.

\subsection{Naturalness}

This condition addresses the artificiality of a setting, such as the nature of the choice task. Our study observes routine officiating decisions with consequences for live matches. Corrections are visible to players, spectators, and television audiences, but the design does not establish how responses vary with the visibility of correction.

A final factor is that Hawk-Eye is nearly perfect at the task we study. Our results are therefore most likely to extend to settings where the reviewing algorithm is highly accurate when it intervenes. Workers may respond differently to less reliable technology because they can reasonably contest its judgment or attribute reversals to algorithmic error.

\subsection{Scaling}

This condition concerns whether responses persist as an intervention scales. During our study period, review was implemented at top tournaments, so umpires and their professional peers were broadly exposed to the same regime. Greater familiarity could normalize correction and weaken its salience, or repeated exposure could strengthen adaptation to the protocol. The larger threshold-shift pattern in the later subperiod is consistent with the latter possibility, but the subperiod analysis cannot separate exposure from other changes over time.

\newpage

\section{Additional Tables and Figures}

\begin{table}[htbp]
\centering
\resizebox{\textwidth}{!}{%
\begin{tabular}{lcccccccc}
\toprule
\textbf{Tournament} & \textbf{Start Month} & \textbf{Category} & \textbf{Court Type} & \multicolumn{4}{c}{\textbf{Number of Matches}} \\
\cmidrule(lr){5-8}
& & & & \textbf{2005} & \textbf{2006} & \textbf{2007} & \textbf{2008} \\
\midrule
Marseille    & Feb & International (250)  & Hard  &        &        &  &  \textcolor{red}{25} \\
Rotterdam    & Feb & International Gold (500)  & Hard  &        &        & \textcolor{red}{15} &  \textcolor{red}{26} \\
Dubai        & Mar & International Gold (500)  & Hard  &        &        &  &  \textcolor{red}{19} \\
Las Vegas     & Mar & International (250)  & Hard  &        &        & &  \textcolor{red}{19} \\
Indian Wells  & Mar & Masters (1000)   & Hard  &  \textcolor{blue}{17}      &  & \textcolor{red}{27} &  \\
Miami        & Mar & Masters (1000)   & Hard  & \textcolor{blue}{15} & \textcolor{red}{18} & \textcolor{red}{26}  &  \\
Queen's Club & Jun & International (250)   & Grass &      \textcolor{blue}{13} & \textcolor{blue}{19} & \textcolor{red}{18}  & \\
UCLA         & Jul & International (250)   & Hard  &        &        & \textcolor{red}{23} & \\
Indianapolis & Jul & International (250)   & Hard  &        & \textcolor{red}{7} & \textcolor{red}{23} &  \\
Washington   & Jul & International (250)   & Hard  &        &        & \textcolor{red}{21} &  \\
Montreal     & Aug & Masters (1000)  & Hard  &  \textcolor{blue}{20} &  & \textcolor{red}{27} &  \\
Toronto      & Aug & Masters (1000)  & Hard  &        &     \textcolor{red}{25}    & &  \\
Cincinnati   & Aug & Masters (1000)  & Hard  &  \textcolor{blue}{11} & \textcolor{red}{19} & \textcolor{red}{24} &  \\
New Haven     & Aug & International (250)   & Hard  &        &        & \textcolor{red}{18} &  \\
Beijing      & Sep & International (250)   & Hard  &        &        & \textcolor{red}{20} &  \\
Kremlin Cup   & Oct & International (250)   & Carpet06 Hard07  &          & \textcolor{red}{8} & \textcolor{red}{15} & \\
Madrid       & Oct & Masters (1000)  & Hard  &             & \textcolor{red}{30} & \textcolor{red}{30} & \\
Basel        & Oct & International (250)   & Hard  &        &        & \textcolor{red}{12} &  \\
Paris Masters& Oct & Masters (1000)  & Carpet06 Hard07 &           & \textcolor{red}{31} & \textcolor{red}{33} &  \\
Shanghai     & Nov & Masters (1000) & Carpet05 Hard06-07      & \textcolor{blue}{14} & \textcolor{red}{15} & \textcolor{red}{15} & \\
\bottomrule
\end{tabular}%
}
\caption{Information on the 35 tournament editions in the consolidated data set. The numbers in each cell indicate the matches available in our data set. The color represents the review regime, with blue indicating tournament editions without Hawk-Eye review and red indicating tournament editions with Hawk-Eye review.}
\label{tab:Tournaments}
\end{table}

\begin{table}[!htbp]
\centering
\small
\resizebox{\textwidth}{!}{%
\begin{tabular}{@{}lcccc@{}}
\toprule
 & Estimate & \shortstack{Clustered\\SE} & \shortstack{Wild-bootstrap\\$p$} & \shortstack{Wild-bootstrap\\95\% CI} \\
\midrule
Closest outside bin $[-20,0)$
    & 0.086 & 0.034 & 0.052 & $[-0.001,\ 0.163]$ \\

First inside bin $[0,20]$
    & -0.041 & 0.034 & 0.256 & $[-0.114,\ 0.040]$ \\

Second inside bin $(20,40]$
    & -0.077 & 0.027 & 0.028 & $[-0.136,\ -0.013]$ \\

\addlinespace
\multicolumn{5}{@{}l}{\textit{Cross-boundary contrasts}} \\

Closest outside $-$ first inside
    & 0.127 & 0.060 & 0.076 & $[-0.019,\ 0.259]$ \\

Closest outside $-$ second inside
    & 0.163 & 0.049 & 0.018 & $[0.048,\ 0.270]$ \\

Closest outside $-$ average of first two inside
    & 0.145 & 0.052 & 0.034 & $[0.017,\ 0.259]$ \\
\bottomrule
\end{tabular}
}
\caption{Wild-Cluster-Bootstrap Inference for the Main Distance-Bin Estimates. The first three rows report estimated changes in the incorrect-call
rate after the introduction of Hawk-Eye review for the indicated 20 mm distance
bins. The remaining rows report contrasts between these estimates. Standard errors
are clustered at the tournament-edition level. Wild-cluster-bootstrap inference is
based on 9,999 replications clustered at the tournament-edition level. For the final
row, the reported estimate compares the closest outside bin with the average of the
first two inside bins. The corresponding wild-bootstrap test is implemented as
$2\alpha_{[-20,0)}-\alpha_{[0,20)}-\alpha_{[20,40)}=0$. Its confidence set is divided
by two so that it is expressed in the same units as the reported contrast. Estimates, standard errors, and interval bounds are in probability units.}
\label{tab:wildbootstrap}
\end{table}

\begin{table}[!htbp]
\centering
\small
\begin{tabular}{@{}lcccc@{}}
\toprule
 & Cross-boundary & Clustered & Wild & Tournament \\
Specification & contrast & SE & $p$ & editions \\
\midrule

\multicolumn{5}{@{}l}{\textit{Panel A. Same-event specification}} \\[2pt]

Full sample
    & 0.145 & 0.052 & 0.034 & 35 \\

Same events + event fixed effects
    & 0.102 & 0.063 & 0.152 & 17 \\

\addlinespace
\multicolumn{5}{@{}l}{\textit{Panel B. Leave-one-event-out estimates}} \\[2pt]

Omitted event
    & Contrast & Clustered SE & \multicolumn{2}{c}{95\% CI} \\
% DAVID VERIFY: Confirm the CI construction and degrees of freedom in Panel B; the wild-bootstrap procedure is documented for the p-values in Panel A.

\cmidrule(lr){4-5}

Canada Open
    & 0.079 & 0.081 & -0.097 & 0.255 \\

Cincinnati
    & 0.110 & 0.064 & -0.030 & 0.250 \\

Indian Wells
    & 0.125 & 0.068 & -0.022 & 0.272 \\

Miami
    & 0.099 & 0.072 & -0.058 & 0.257 \\

Queen's Club
    & 0.103 & 0.059 & -0.026 & 0.232 \\

Shanghai
    & 0.077 & 0.067 & -0.069 & 0.222 \\

\bottomrule
\end{tabular}

\caption{Same-Event and Leave-One-Event-Out Robustness. The cross-boundary contrast is defined as the estimated change in the incorrect-call rate for the closest outside bin, $[-20,0)$, minus the average estimated change for the first two inside bins, $[0,20)$ and $[20,40)$. The same-event specification restricts the sample to named events observed under both review regimes and includes named-event fixed effects. Toronto and Montreal are treated as a single event, the Canada Open, because the tournament alternates between the two host cities. Standard errors are clustered at the tournament-edition level. The $p$-values in Panel A use wild-cluster-bootstrap inference with 9,999 replications clustered at the tournament-edition level. Panel B re-estimates the same-event specification
after omitting each named event in turn. The leave-one-event-out estimates are
reported to assess sensitivity to individual events rather than for separate
hypothesis testing. Estimates, standard errors, and interval bounds are in probability units.}
\label{tab:sameevent}
\end{table}

\begin{table}[!htbp]
\centering
\small
\begin{tabular}{@{}lccc@{}}
\toprule
Specification & \shortstack{Cross-boundary\\contrast} & \shortstack{Clustered\\SE} & 95\% CI \\
% DAVID VERIFY: Confirm the CI method, critical value/degrees of freedom, and any finite-sample adjustment. The rounded bounds appear consistent with t(34), but this is not established by the available output. Preserve the reported numbers until confirmed.
\midrule
Baseline
    & 0.145 & 0.052 & $[0.040,\ 0.251]$ \\

+ Serve/non-serve control
    & 0.144 & 0.052 & $[0.038,\ 0.250]$ \\

+ Flexible speed controls
    & 0.144 & 0.052 & $[0.039,\ 0.250]$ \\

+ Hitter and receiver fixed effects
    & 0.136 & 0.053 & $[0.028,\ 0.244]$ \\

\bottomrule
\end{tabular}
\caption{Robustness to Shot and Player Composition. The cross-boundary contrast is defined as the estimated change in the incorrect-call rate for the closest outside bin, $[-20,0)$, minus the average estimated change for the first two inside bins, $[0,20)$ and $[20,40)$. The baseline specification includes 20 mm signed-distance-bin fixed effects, an indicator for ball speed above the sample median, indicators for the server's point score, game number, set number, round, and tournament tier, and a tie-break indicator. Controls are added sequentially across rows: the second row adds serve status, the third replaces the above-median-speed indicator with flexible controls for ball speed, and the fourth additionally includes hitter and receiver player fixed effects. Standard errors are clustered at the tournament-edition level. Estimates, standard errors, and interval bounds are in probability units.}
\label{tab:composition}
\end{table}

\begin{table}[!htbp]
\centering
\small
\begin{tabular}{@{}lcccc@{}}
\toprule
& \multicolumn{2}{c}{Joint model}
& \multicolumn{2}{c}{Reversals only} \\
\cmidrule(lr){2-3}\cmidrule(l){4-5}
Prior reversals & Asymmetry & $p$-value & Asymmetry & $p$-value \\
\midrule
\multicolumn{5}{@{}l}{\textit{Panel A. Flexible specification}} \\[3pt]

One
    & 0.115 & 0.061 & 0.125 & 0.040 \\
    & (0.061) & & (0.061) & \\[3pt]

Two
    & 0.144 & 0.167 & 0.178 & 0.081 \\
    & (0.104) & & (0.102) & \\[3pt]

Three or more
    & 0.107 & 0.504 & 0.156 & 0.324 \\
    & (0.159) & & (0.158) & \\

\addlinespace
Joint symmetry test ($p$-value)
    & \multicolumn{2}{c}{0.254}
    & \multicolumn{2}{c}{0.151} \\

\addlinespace
\multicolumn{5}{@{}l}{\textit{Panel B. Linear specification}} \\[3pt]

Per additional reversal
    & 0.089 & 0.032 & 0.103 & 0.011 \\
    & (0.042) & & (0.041) & \\

\bottomrule
\end{tabular}
\caption{Asymmetry in Responses to Prior Reversals. Each asymmetry
estimate is the sum of the coefficients on the corresponding out-to-in
and in-to-out reversal variables. A positive estimate indicates a larger
response in absolute magnitude following an out-to-in reversal. The joint
model includes histories of both reversal types and both types of upheld
challenges; its flexible specification is the one used in Figure~3.
The reversals-only model excludes upheld-challenge histories.
Panel A allows separate effects after one, two, and three or more prior
reversals. The joint symmetry test evaluates whether all three
asymmetries equal zero ($F(3,567)=1.36$ for the joint model and
$F(3,567)=1.78$ for the reversals-only model). Panel B enters reversal
counts linearly and reports asymmetry per additional reversal.
Match-clustered standard errors appear in parentheses. Individual
$p$-values are two-sided tests of zero asymmetry. Estimates and standard
errors are in probability units.}
\label{tab:reversal_asymmetry}
\end{table}

\begin{table}[!htbp]
\centering
\small
\begin{tabular}{@{}lccc@{}}
\toprule
Comparison & Estimate & Clustered SE & $p$-value \\
\midrule
IN$\rightarrow$OUT reversal minus upheld IN call
    & -0.067 & 0.038 & 0.078 \\

OUT$\rightarrow$IN reversal minus upheld OUT call
    & 0.115 & 0.040 & 0.004 \\

\addlinespace
Joint equality with unsuccessful challenges
    & --- & --- & 0.004 \\

Difference in absolute incremental effects
    & 0.049 & 0.053 & 0.363 \\

\bottomrule
\end{tabular}
\caption{The table compares estimated associations of subsequent calls with
prior reversals and with unsuccessful challenges involving the same initial-call direction. A negative
estimate in the first row indicates an additional shift toward calling balls out
following an IN-to-OUT reversal. A positive estimate in the second row indicates
an additional shift toward calling balls in following an OUT-to-IN reversal.
All four cumulative feedback variables are included simultaneously in the same
within-match specification used in Section~\ref{sec:within_match}. Standard errors are clustered at the match level. The joint test evaluates whether the two reversal coefficients equal the corresponding coefficients on unsuccessful challenges. Estimates and standard errors are in probability units. The final row tests whether the incremental associations are equal in absolute magnitude.}
\label{tab:reversal_upheld}
\end{table}

\begin{table}%[htbp]
\centering
\def\sym#1{\ifmmode^{#1}\else\(^{#1}\)\fi}
\small
\scalebox{.9}{
\begin{tabular}{l*{4}{c}}
\hline\hline
                &\multicolumn{1}{c}{(1)}&\multicolumn{1}{c}{(2)}&\multicolumn{1}{c}{(3)}  &\multicolumn{1}{c}{(4)} \\
                &\multicolumn{4}{c}{Incorrect call}\\
\hline
PostHK &  -0.014\sym{**} &  -0.011\sym{} & -0.011\sym{} & -0.011\sym{} \\
                & (0.007)   & (0.007)   & (0.007)   & (0.008)    \\
[1em]
\shortstack[l]{Call and point\\controls} & & X & X & X \\
[.5em]
Match controls       &    &    &   X & X  \\
[.5em]
% DAVID VERIFY: Identify the SE estimator (conventional OLS or heteroskedasticity-robust) and test reference distribution in columns 1--3; blank cluster entries do not establish the method.
Cluster level & & & & \shortstack{Tournament\\edition} \\
\hline 
N              & 16,812  &  16,812  &  16,335  &  16,335  \\
[.5em]
Baseline mean           &  .139  &  .139   &  .136  &  .136 \\
\hline\hline
\multicolumn{5}{l}{\footnotesize Standard errors in parentheses}\\
\multicolumn{5}{l}{\footnotesize \sym{*} \(p<.10\), \sym{**} \(p<.05\), \sym{***} \(p<.01\)}\\
\end{tabular}
}
% DAVID VERIFY: Identify the precise reason for the 477 exclusions when match controls are included (263 serves and 214 non-serves). The reported counts establish the sample reduction, not which variables or restrictions cause it.
\caption{OLS regressions of umpire mistakes for balls bouncing within 100 mm of the line. $PostHK=1$ if Hawk-Eye review is active. Call and point controls include 20 mm signed-distance-bin fixed effects, an indicator for ball speed above the sample median, indicators for the server’s point score, and a tie-break indicator. Match controls include indicators for game number, set number, round, and tournament tier. Including match controls reduces the sample from 16,812 to 16,335 calls, excluding 477 observations (263 serves and 214 non-serves) with missing values for the game or set variables. Coefficients, standard errors, and baseline means are in probability units (0.01 is one percentage point).}
\label{tab:regressions_pooled}
\end{table}

\FloatBarrier

\begin{figure}[htbp]
\centering
\includegraphics[width=5in]{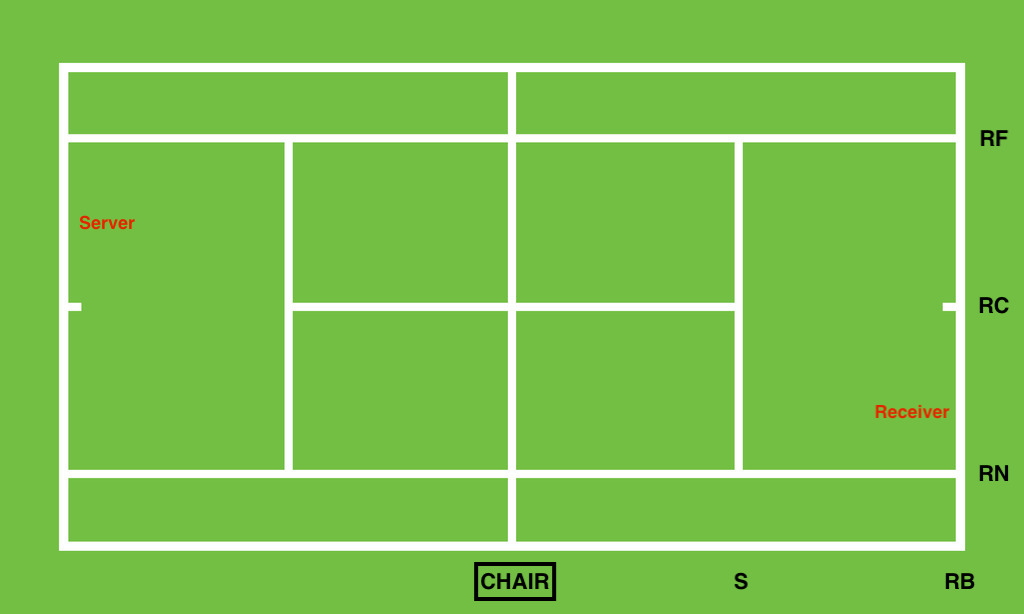}
\caption{The location of the umpiring crew: S = the serve-line umpire, RB = the right baseline umpire, RN = the right near long-line umpire, RC = the right center-line umpire, RF = the right far-side long-line umpire. The left side of the court has corresponding line umpires, except for the serve-line umpire, who moves to the left side when the player on the right side is serving.}
\label{fig:Umpires}
\end{figure}

\begin{figure}[!htbp]
\centering
\includegraphics[width=5.5in]{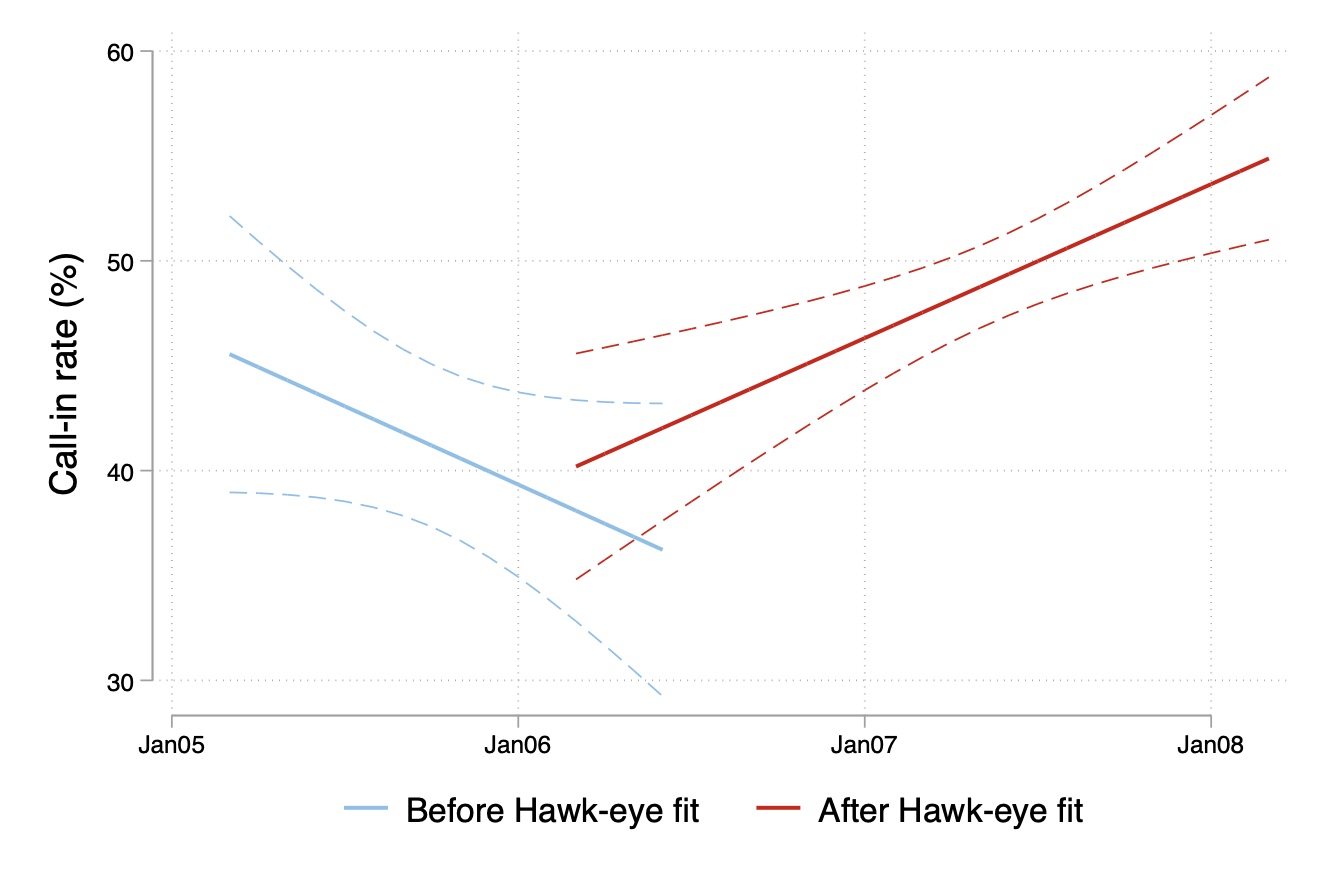}
\caption{Call-in rates before and after the introduction of Hawk-Eye review for balls landing within 20 mm of the line, regardless of which side of the line they actually bounced on. The solid lines report fitted values from a linear probability model in which the probability of calling a ball in is allowed to vary linearly with month, separately before and after Hawk-Eye review. The dashed lines indicate 95\% confidence intervals. The regression controls for game number, set number, the server's point score, tie-break status, round, tournament tier, and shot speed. Standard errors are clustered at the match level.}
\label{fig:In}
\end{figure}

\begin{figure}
\centering
\begin{subfigure}[b]{.45\textwidth}
\centering
\includegraphics[width=2.9in]{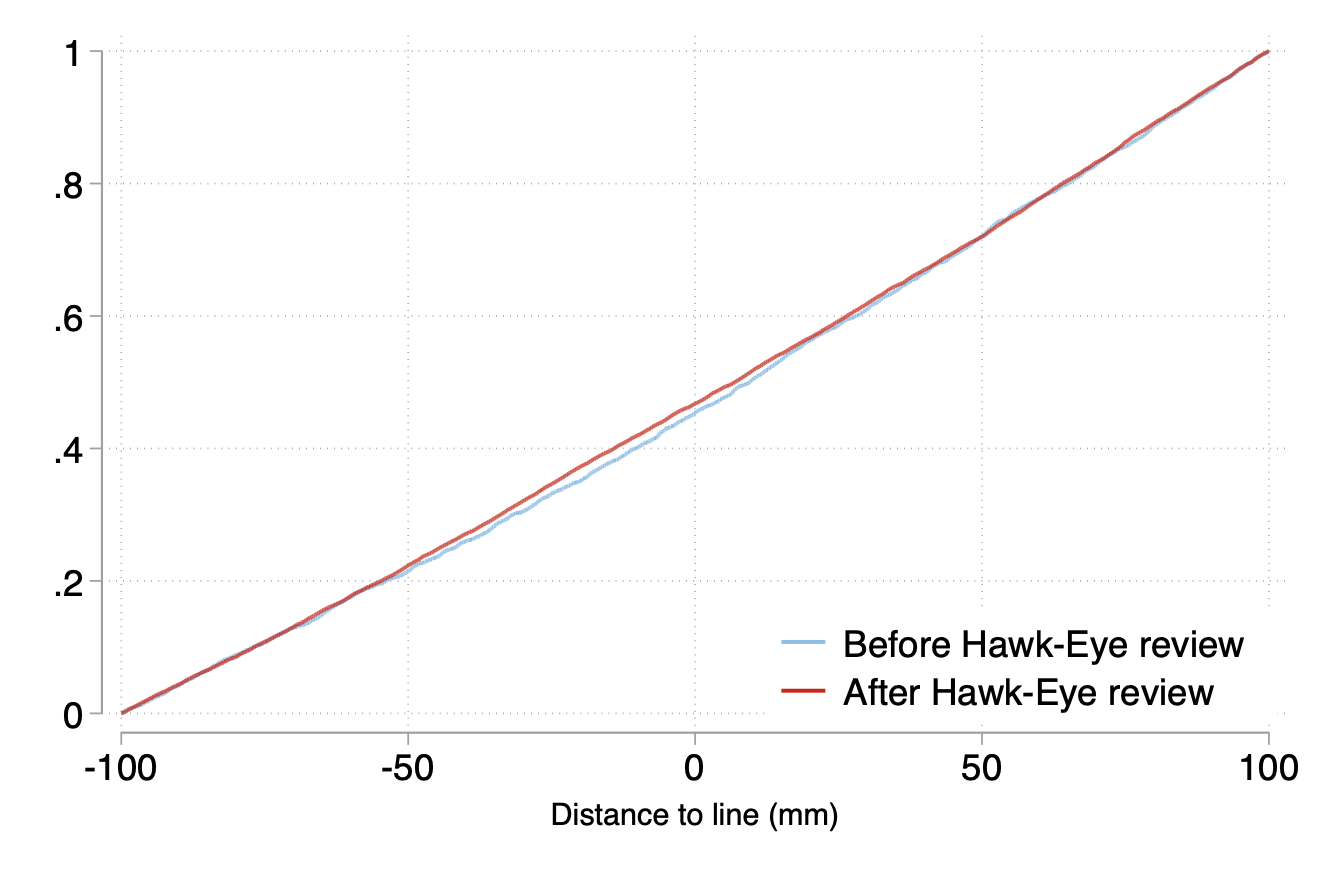}
\caption{Distance to line CDF.}
\end{subfigure}%
\begin{subfigure}[b]{.49\textwidth}
\centering
\includegraphics[width=2.95in]{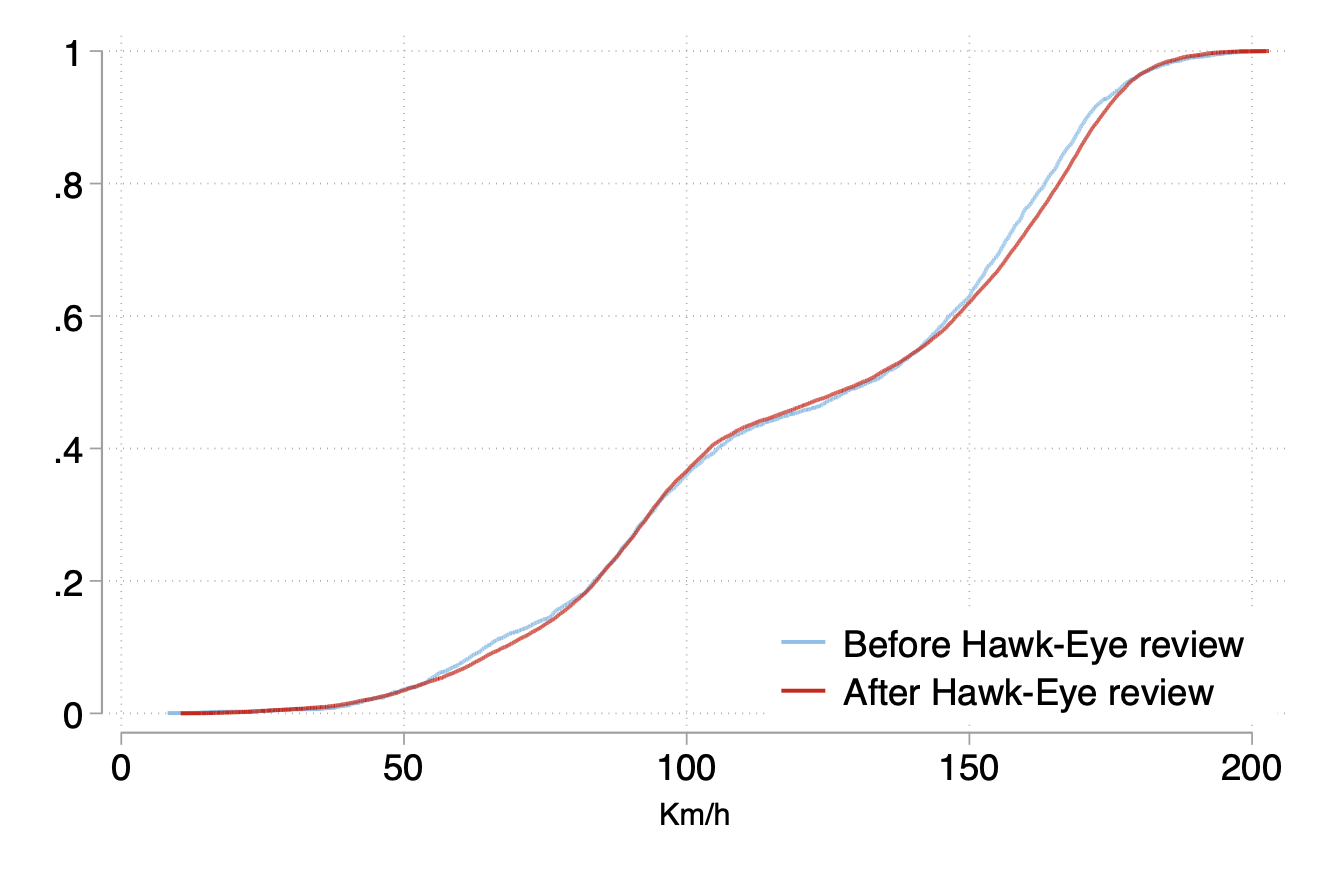}
\caption{Speed CDF.}
\end{subfigure}
\caption{Empirical distributions of signed distance from the line and ball speed before and after Hawk-Eye review.}
\label{fig:PlayerStrokes}
\end{figure}

\begin{figure}
\centering
\includegraphics[width=5.5in]{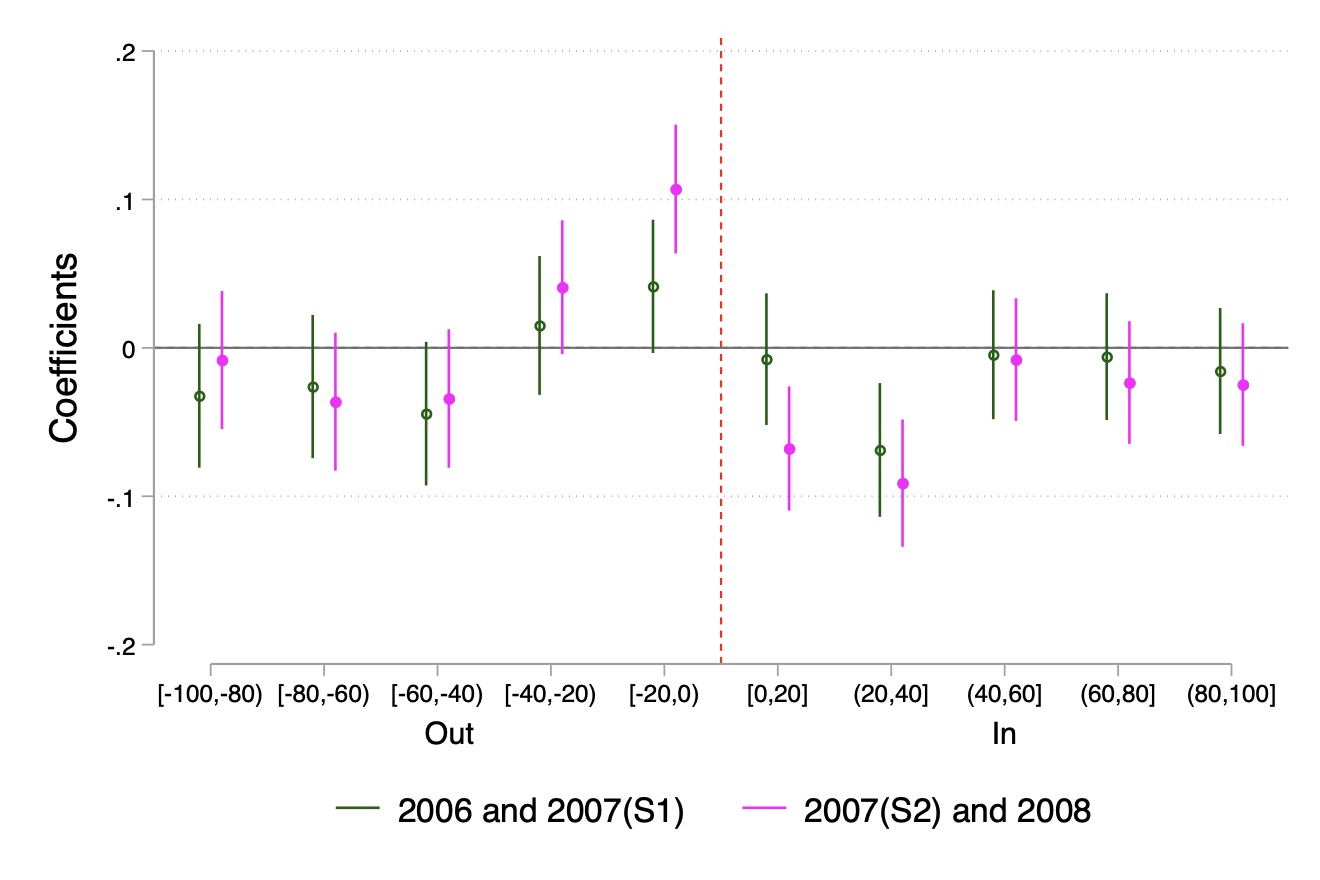}
\caption{Subperiod analysis of changes in the incorrect-call rate after the introduction of Hawk-Eye review, by proximity to the line. Matches with review are grouped into tournaments from 2006 through the first half of 2007 and tournaments from the second half of 2007 through 2008. Each dot represents the coefficient on the interaction between a distance-bin indicator and $PostHK$.}
\label{fig:ES}
\end{figure}

\begin{figure}[!htbp]
\centering
\includegraphics[width=5.5in]{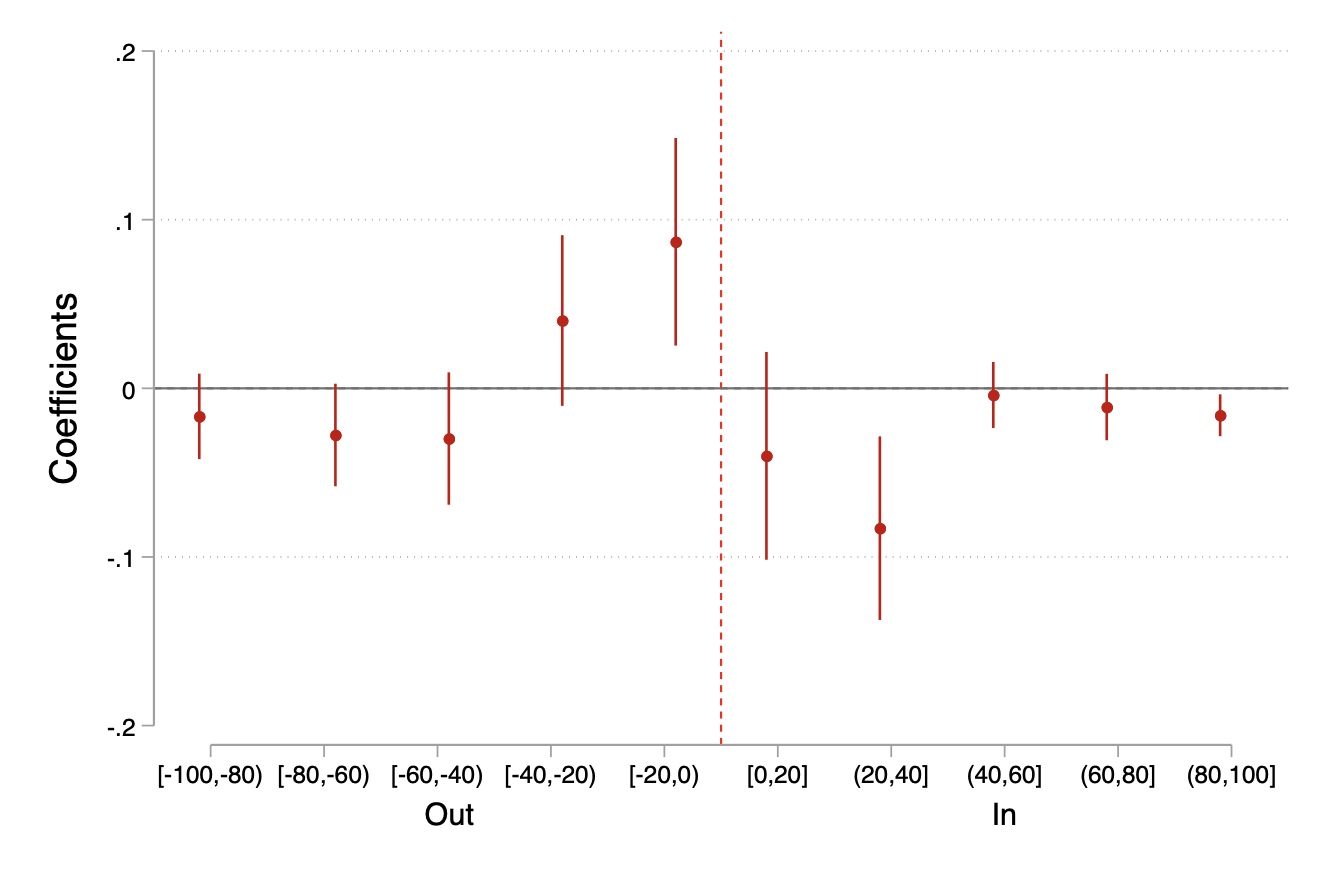}
\caption{An analog of Figure~\ref{fig:HE}, using a one-treatment-at-a-time approach. We estimate the effect for each distance bin separately to avoid contamination across bins.}
\label{fig:Contamination}
\end{figure}

% \begin{figure}[!htbp]
% \centering
% \begin{subfigure}[b]{.45\textwidth}
% \centering
% \includegraphics[width=\linewidth]{Figures/in_serve.jpg}
% \caption{Serves.}
% \end{subfigure}
% \begin{subfigure}[b]{.45\textwidth}
% \centering
% \includegraphics[width=\linewidth]{Figures/in_nonserve.jpg}
% \caption{Non-serves.}
% \end{subfigure}
% \caption{An analog of Figure \ref{fig:In}, separating serves and non-serves.}
% \label{fig:InByserve}
% \end{figure}

% \begin{figure}[!htbp]
% \centering
% \includegraphics[width=5.5in]{Figures/Psychometric.jpg}
% \caption{The dots represent the average rate at which umpires call a ball ``in'' when it bounces within a 10 mm radius (e.g., at x=0 is the average rate for balls bouncing between -10 mm and +10 mm). The psychometric curves were approximated using locally weighted scatterplot smoothing.}
% \label{fig:Psychometric}
% \end{figure}

%\begin{figure}[!htbp]
%\centering
%\begin{subfigure}[b]{.45\textwidth}
%\centering
%\includegraphics[width=3in]{Figures/HE_serve.jpg}
%\caption{Serves.}
%\end{subfigure}
%\begin{subfigure}[b]{.45\textwidth}
%\centering
%\includegraphics[width=3in]{Figures/HE_nonserve.jpg}
%\caption{Non-serves.}
%\end{subfigure}
%\caption{An analog of Figure \ref{fig:HE}, separating serves and non-serves.}
%\label{fig:HEByserve}
%\end{figure}

\begin{figure}[!htbp]
\centering
\includegraphics[width=5.5in]{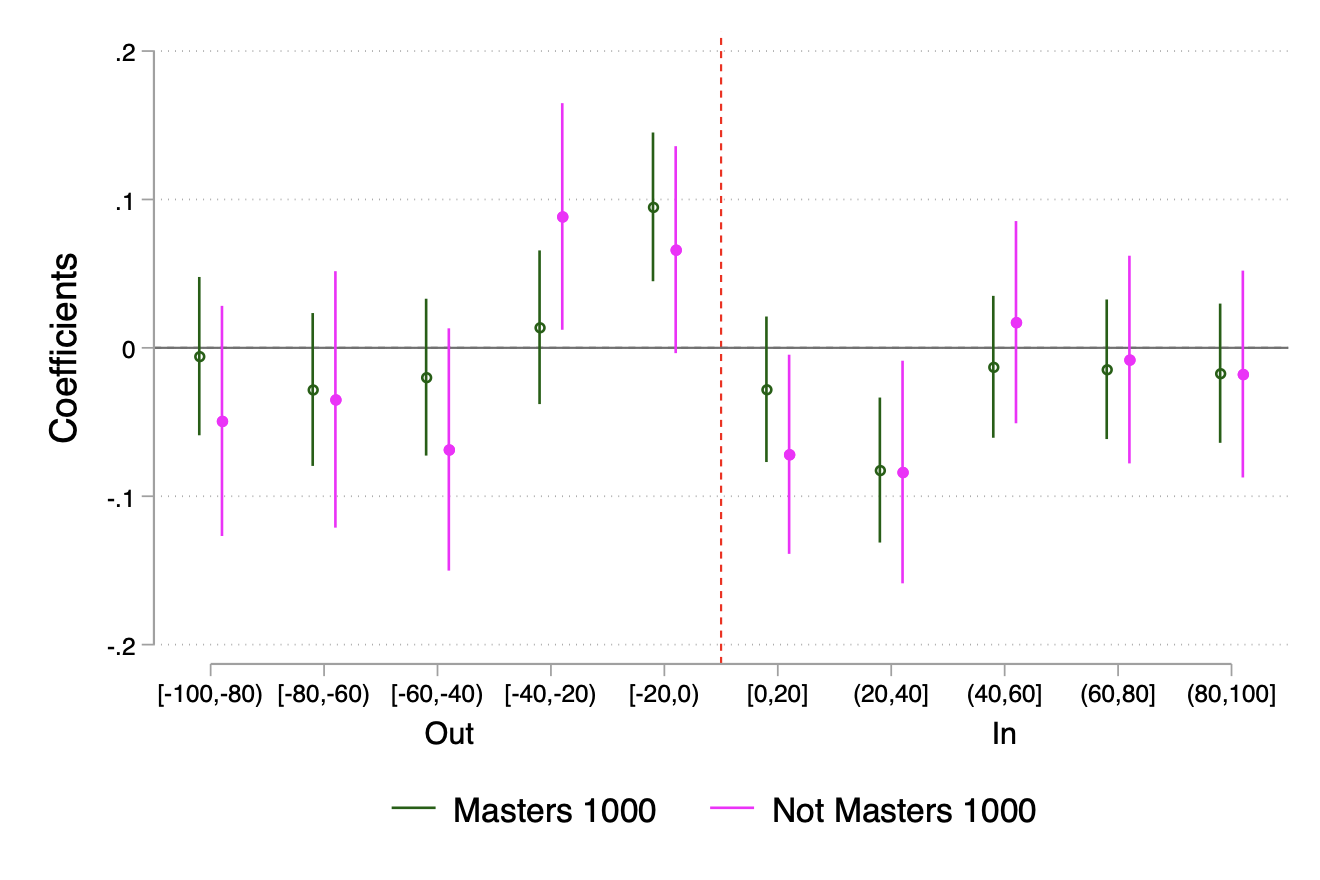}
\caption{Estimated change in umpire performance after the introduction of Hawk-Eye review, across 20 mm distance bins and by tournament category. Matches are grouped into higher-tier tournaments (Masters 1000) and lower-tier tournaments (ATP 500 and 250). Each dot represents the coefficient on the interaction between a distance-bin indicator and $PostHK$, estimated separately by tournament category.}
\label{fig:HE_Cat}
\end{figure}

\begin{figure}[!htbp]
\centering
\includegraphics[width=5in]{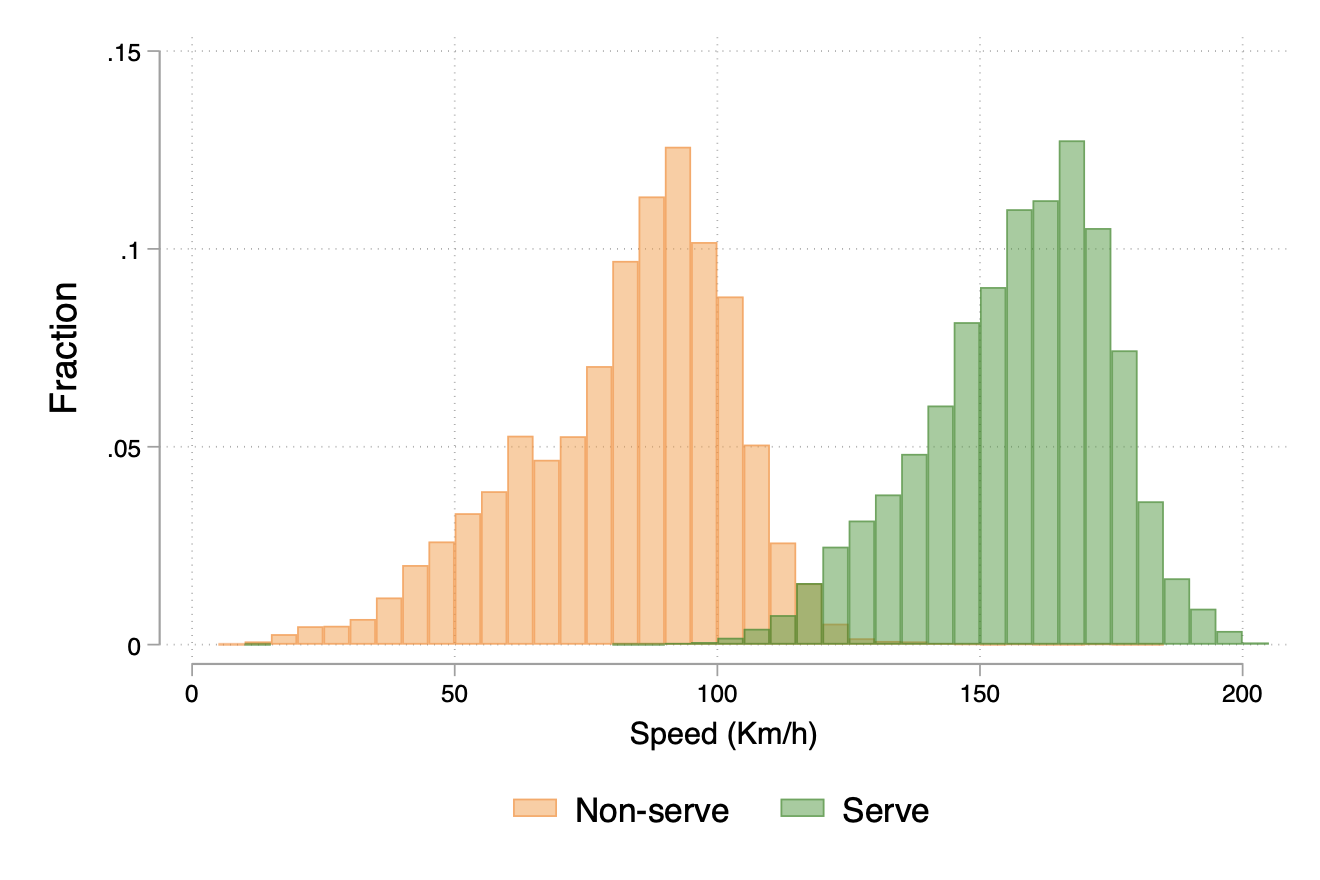}
\caption{Speed distribution for all the balls that bounced within 100 mm of the line, separated by non-serves and serves.}
\label{fig:Speed}
\end{figure}

\begin{figure}
\centering
\begin{subfigure}[b]{.45\textwidth}
\centering
\includegraphics[width=2.8in]{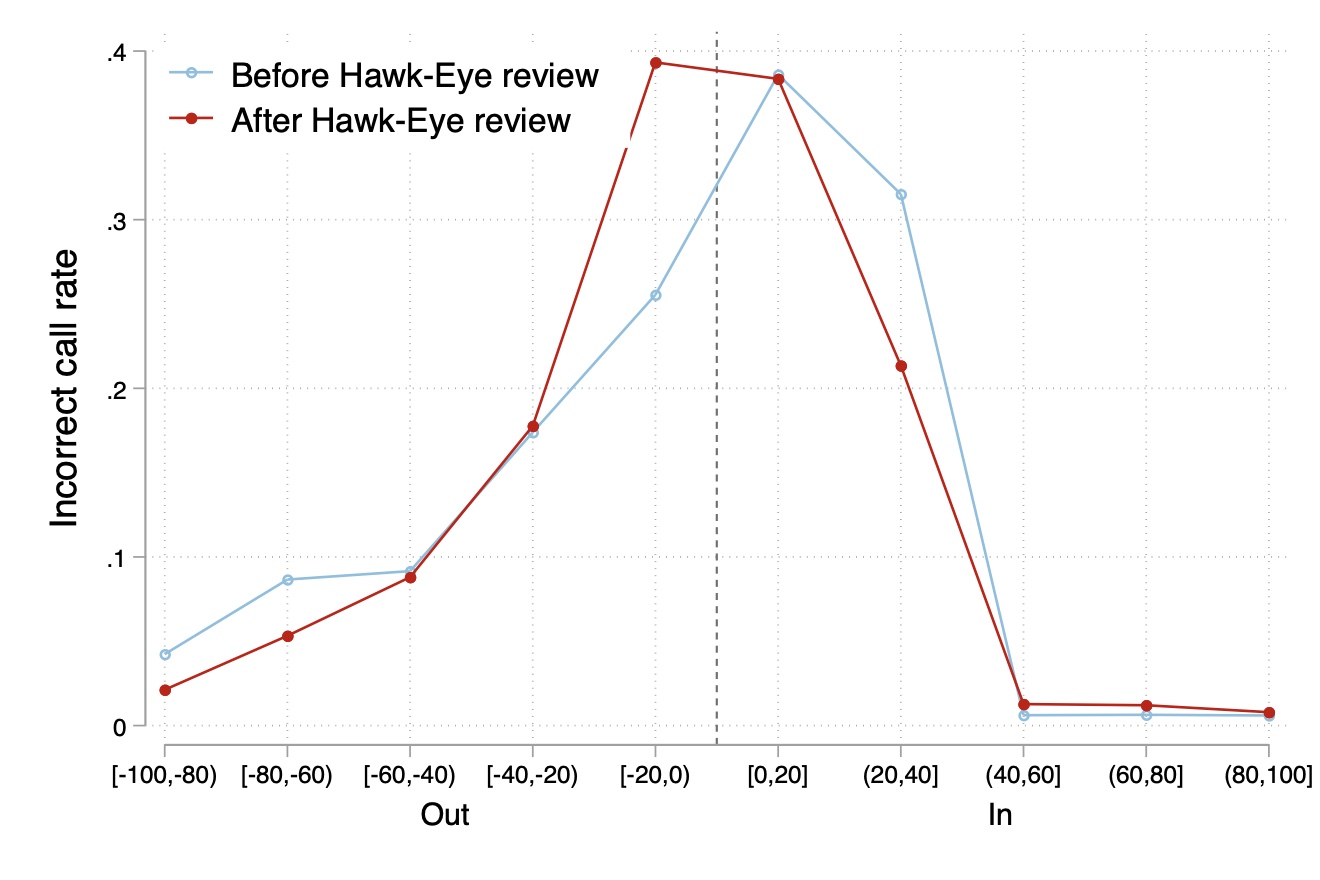}
\caption{Serves.}
\end{subfigure}
\begin{subfigure}[b]{.45\textwidth}
\centering
\includegraphics[width=2.8in]{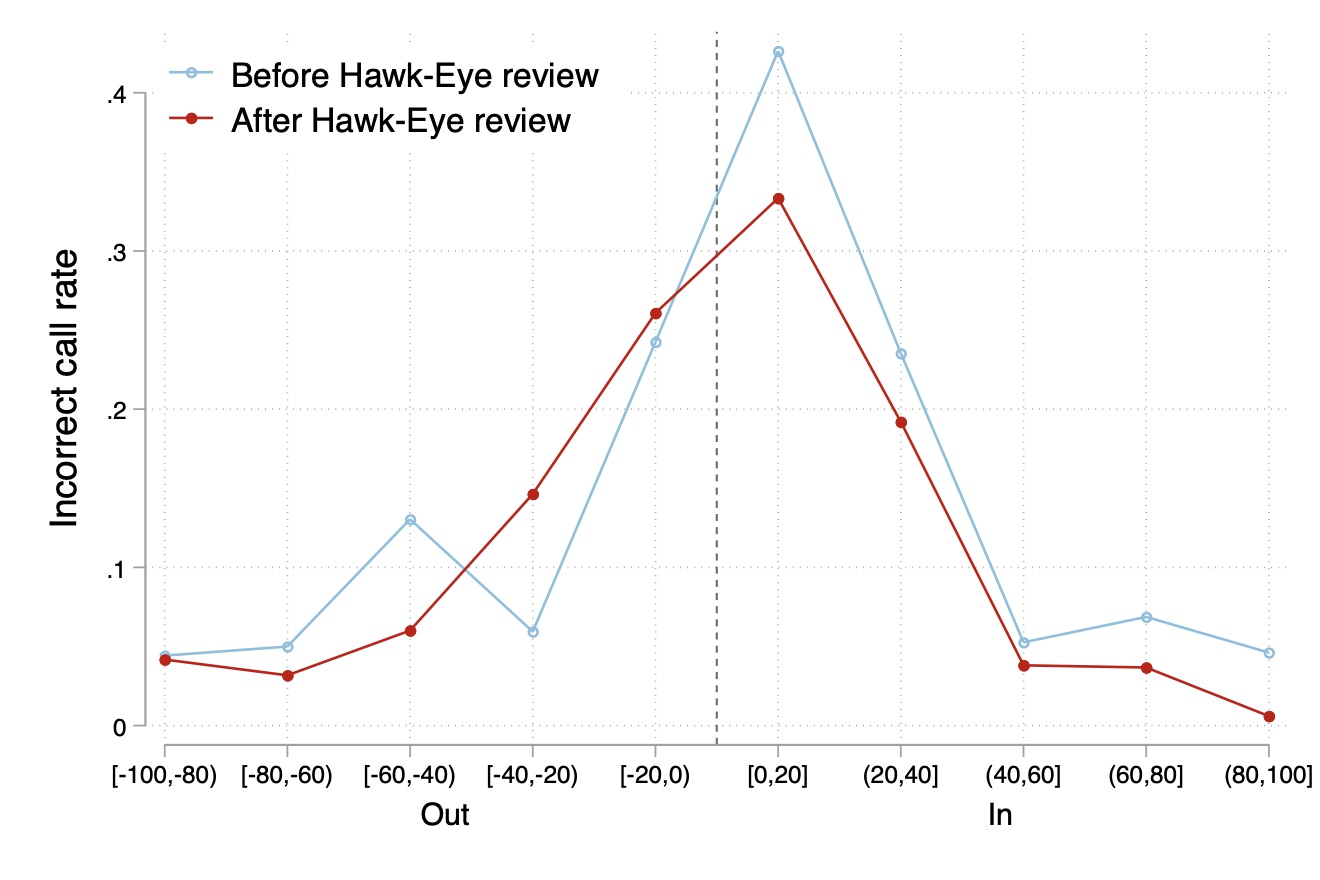}
\caption{Non-serves.}
\end{subfigure}
\caption{Incorrect call rates by proximity to the line (separately for serves and non-serves). Each dot is the rate of incorrect calls for a bin of 20 mm, and dots to the left of the dashed line represent bins out of bounds, and those to the right of the dashed line represent bins in bounds.}
\label{fig:Mistakes_byserve}
\end{figure}

\FloatBarrier
\section{Pooled Call and Correction Statistics}\label{app:structural_moments}

This appendix reports the aggregate inputs needed to reconstruct initial-call probabilities and correction exposure for balls landing within 100 mm from the line as reported in Table \ref{tab:review_error_paths}. Every call receives equal weight, and calls landing farther than 100 mm from the line are excluded. Table \ref{tab:pooled_call_counts} crosses the ball's actual location with the initial human call, before any algorithmic correction. The final column counts initially incorrect calls corrected by review, not all challenges or reviews.

% SOURCE: David Almog, "Simple stats needed," October 15, 2026, PDF p. 2.
% True-in totals: 1433 without review, 7554 with review.
% True-out totals: 1189 without review, 6636 with review.
% Correct initial calls equal the state total minus that state's initial errors.
% Corrected mistakes equal initial minus remaining mistakes under A_j=0.
\begin{table}[!htbp]
\centering
\small
\setlength{\tabcolsep}{5pt}
\begin{tabular}{@{}llrrr@{}}
\toprule
Actual location & Initial call & \shortstack{Without\\review} & \shortstack{With\\review} & \shortstack{Corrected\\with review}\\
\midrule
In & In & 1,216 & 6,664 & 0\\
In & Out & 217 & 890 & 192\\
Out & In & 147 & 879 & 297\\
Out & Out & 1,042 & 5,757 & 0\\
\midrule
Total & & 2,622 & 14,190 & 489\\
\bottomrule
\end{tabular}
\caption{Initial calls by actual location and subsequent correction.}
\label{tab:pooled_call_counts}
\par\smallskip
\begin{minipage}{\textwidth}
\footnotesize\textit{Notes:} Entries are counts, not percentages. Actual location is determined by Hawk-Eye. Corrections are the reduction in mistakes between the initial and final line classifications. The zero entries for initially correct calls indicate that there is no initial error to correct, not that such calls are never reviewed. The table does not report reviews that uphold a call or their duration. Without algorithmic review, no initial mistakes are corrected by that system.
\end{minipage}
\end{table}

Table \ref{tab:pooled_call_moments} reports the resulting conditional probabilities. Let $r=0$ denote no review and $r=1$ review. The quantities $p_r$ and $q_r$ condition on the initial call, whereas $m_r$ is the actual in share and $\lambda_r$ is the initial call-in share. For example, $p_0=1{,}216/(1{,}216+147)$ and $q_0=217/(217+1{,}042)$. As in Section \ref{sec:review_returns}, correction fractions are indexed by the direction of the incorrect initial call. Thus, $\eta_{\mathrm{in}}=297/879$ is the fraction of mistaken in calls corrected, conditional on the ball being out and initially called in. Likewise, $\eta_{\mathrm{out}}=192/890$ is the fraction of mistaken out calls corrected, conditional on the ball being in and initially called out.

\begin{table}[!htbp]
\centering
\small
\setlength{\tabcolsep}{5pt}
\renewcommand{\arraystretch}{1.16}
\begin{tabular}{@{}lrr@{}}
\toprule
Quantity and conditioning & Without review & With review\\
\midrule
$p_r=\Pr(\text{ball in}\mid\text{initial call in})$ & 0.89214967 & 0.88346812\\
$q_r=\Pr(\text{ball in}\mid\text{initial call out})$ & 0.17235902 & 0.13389499\\
$m_r=\Pr(\text{ball in})$ & 0.54652937 & 0.53234672\\
$\lambda_r=\Pr(\text{initial call in})$ & 0.51983219 & 0.53157153\\
$\eta_{\mathrm{in}}=\Pr(\text{corrected}\mid\text{ball out, initial call in})$ & 0 & 0.33788396\\
$\eta_{\mathrm{out}}=\Pr(\text{corrected}\mid\text{ball in, initial call out})$ & 0 & 0.21573034\\
\bottomrule
\end{tabular}
\caption{Pooled probabilities implied by Table \ref{tab:pooled_call_counts}.}
\label{tab:pooled_call_moments}
\par\smallskip
\begin{minipage}{\textwidth}
\footnotesize\textit{Notes:} Probabilities are rounded to eight decimal places. Calculations should use the integer counts in Table \ref{tab:pooled_call_counts}, rather than rounded entries. Under the ground-truth benchmark, correction exposure measures successful correction of an initially incorrect line classification, not the probability that an arbitrary call is challenged.
\end{minipage}
\end{table}

\FloatBarrier
The integer counts reproduce the pooled moments without access to individual call records, but do not preserve dependence within matches or tournament editions. Clustered uncertainty therefore requires these counts separately by resampling unit, or the corresponding bootstrap moment draws. In particular, a tournament-edition bootstrap requires the four initial state--call counts and two corrected-error counts for each edition, together with its review status, rather than pooled totals alone.
\end{document}